\documentclass[sigconf,review，screen]{acmart}
\usepackage{bbding}
\usepackage{caption}
\usepackage{subcaption}
\usepackage{fontawesome}
\usepackage{CJKutf8}
\AtBeginDocument{%
  }

\copyrightyear{2025}
\acmYear{2025}
\setcopyright{acmlicensed}
\acmConference[MM '25] {Proceedings of the 33rd ACM International Conference on Multimedia}{October 27--31, 2025}{Dublin, Ireland.}
\acmBooktitle{Proceedings of the 33rd ACM International Conference on Multimedia (MM '25), October 27--31, 2025, Dublin, Ireland}
\acmISBN{979-8-4007-2035-2/2025/10}

\acmConference[Conference acronym 'XX]{Make sure to enter the correct
  conference title from your rights confirmation email}{October 27--31,
  2025}{Dublin, Ireland}
\acmDOI{10.1145/XXXXXX.XXXXXX}
\usepackage{multirow}

\acmSubmissionID{39}

\begin{document}
\begin{CJK}{UTF8}{gbsn}
\title{EventVAD: Training-Free Event-Aware Video Anomaly Detection}

% \author{Yihua Shao$^1$, \quad Haojin He$^{1,2}$, \quad Sijie Li$^{1,3}$, \quad Siyu Chen$^4$, \quad Xinwei Long$^1$, \quad Fanhu Zeng$^1$, \quad Yuxuan Fan$^1$, \quad Muyang Zhang$^1$, \quad Ziyang Yan$^1$, \quad Ao Ma$^1$, \quad Xiaochen Wang$^1$, \quad Hao Tang$^1$, \quad Yan Wang$^1$, \quad Shuyan Li$^1$}
% \affiliation{%
% 	\institution{$^1$Department of Information Engineering and Computer Science, University of Trento, Trento, Italy \\
% 		$^2$Computer Vision Laboratory, \'Ecole Polytechnique F\'ed\'erale de Lausanne, Lausanne, Switzerland \\
% 		$^3$Department of Engineering Science, University of Oxford, Oxford, United Kingdom \\
% 		$^4$Department of Computer Science, Texas State University, San Marcos, USA}

% }

\author{Yihua Shao}

\affiliation{%
  \institution{Peking University}
  \country{China}}
\email{yihuajerry@gmail.com}

\author{Haojin He}
\authornote{Haojin He is equal contribution with Yihua Shao.}
\affiliation{%
  \institution{Guangdong University of Technology}
  \country{China}}
\email{3122009330@mail2.gdut.edu.cn}

\author{Sijie Li}
\affiliation{%
  \institution{The University of Sheffield}
  \country{The United Kingdom}}
  \email{sli256@sheffield.ac.uk}

\author{Siyu Chen}
\affiliation{%
 \institution{University of Science and Technology Beijing}
 \country{China}}
 \email{siyuchen200311@163.com}

\author{Xinwei Long}
\authornote{Xinwei Long is the minor corresponding author.}
\affiliation{%
  \institution{Tsinghua University}
  \country{China}}
\email{longxw22@mails.tsinghua.edu.cn}

\author{Fanhu Zeng}
\affiliation{%
  \institution{Tsinghua University}
  \country{China}
}
\email{zengfanhu2022@ia.ac.cn}

\author{Yuxuan Fan}
\affiliation{%
  \institution{The Hong Kong University of Science and Technology (Guangzhou)}
  \country{China}
}
\email{orionisfan@outlook.com}

\author{Muyang Zhang}
\affiliation{%
  \institution{Nanjing University}
  \country{China}}
\email{zmy417419@163.com}

\author{Ziyang Yan}
\affiliation{%
  \institution{University of Trento}
  \country{Italy}}
\email{ziyang.yan@unitn.it}

\author{Ao Ma}
\authornote{Ao Ma is the project leader.}
\affiliation{%
  \institution{JD.com}
  \country{China}}
\email{maaoaoma@126.com}

\author{Xiaochen Wang}
\affiliation{%
  \institution{University of Science and Technology Beijing}
  \country{China}
}
\email{wangxiaochen@ustb.edu.cn}

\author{Hao Tang}
\affiliation{%
  \institution{Peking University}
  \country{China}}
\email{haotang@pku.edu.cn}

\author{Yan Wang}
\affiliation{%
  \institution{Tsinghua University}
  \country{China}
}
\email{wangyan@air.tsinghua.edu.cn}

\author{Shuyan Li}
\authornote{Shuyan Li is the major corresponding author.}
\affiliation{%
  \institution{Queen's University Belfast}
  \country{The United Kingdom}}
\email{li-sy16@tsinghua.org.cn}

\renewcommand{\shortauthors}{Trovato et al.}

\begin{abstract}
Video Anomaly Detection~(VAD) focuses on identifying anomalies within videos. 
% However, the trained method is relaying on training data. 
% When faced with an unseen anomaly, it need to collect data and retrain the model to maintain optimal performance. 
% While training-free approaches enhance generalization, it is difficult to manage the numerous events and fine-grained anomaly in long videos. 
Supervised methods require an amount of in-domain training data and often struggle to generalize to unseen anomalies.
In contrast, training-free methods leverage the intrinsic world knowledge of large language models (LLMs) to detect anomalies but face challenges in localizing fine-grained visual transitions and diverse events.
Therefore, we propose \textbf{EventVAD}, an event-aware video anomaly detection framework that combines tailored dynamic graph architectures and multimodal LLMs to perform fine-grained temporal-event reasoning.
% Therefore, we propose \textbf{EventVAD}, an event-driven video anomaly detection framework that achieves event-aware modeling through a synergistic integration of Dynamic Graph architectures and multimodal large language models (MLLMs).
Specifically,
EventVAD first employs dynamic spatiotemporal graph modeling with time-decay constraints to capture event-aware video features.
Then, it performs adaptive noise filtering and uses signal ratio thresholding to detect event boundaries via unsupervised statistical features.
% Then, it utilizes adaptive filtering to mitigate noises and robust signal ratio thresholding to detect event boundaries through statistical features without any supervision.
% identify event transitions, which
% identify event transitions via robust signal ratio thresholding, which detects event boundaries through statistical features without any supervised signals.
% enabling statistical boundary detection.
% Then, it utilizes adaptive filtering and statistical boundary detection to parse videos into distinct events.
% The statistical boundary detection module reduces the complexity of processing long videos for MLLMs and improves their temporal reasoning through event consistency.
Finally, it utilizes a hierarchical prompting strategy to guide MLLMs in performing reasoning and making final decisions.
%reason over event sequences before making the final anomaly decision.
% EventVAD implements event-aware video parsing by employing dynamic spatiotemporal graph modeling with time-decay constraints. 
% By segmenting videos into distinct events, EventVAD can reduce the processing difficulty of long videos for MLLMs while enhancing their temporal understanding through event consistency. 
% % Additionally, it utilizes adaptive filtering and statistical boundary detection for orthogonal feature propagation. 
% This framework enables accurate event segmentation before MLLM-based anomaly scoring. 
We conducted extensive experiments on the UCF-Crime and XD-Violence datasets.
The results demonstrate that EventVAD with a 7B MLLM achieves \textit{state-of-the-art (SOTA)} in training-free settings, outperforming strong baselines that use 7B or larger MLLMs. The code is available at \textbf{\href{https://github.com/YihuaJerry/EventVAD}{https://github.com/YihuaJerry/EventVAD}}.
% while requiring only a 7B foundational model compared with other training-free method's 13b foundational model.
\end{abstract}



\received{20 February 2007}
\received[revised]{12 March 2009}
\received[accepted]{5 June 2009}

\maketitle

\begin{figure}[htbp]
  \centering
  \subfloat[Previous Training-Free Video Anomaly Detection]{
  \includegraphics[width=0.45\textwidth]{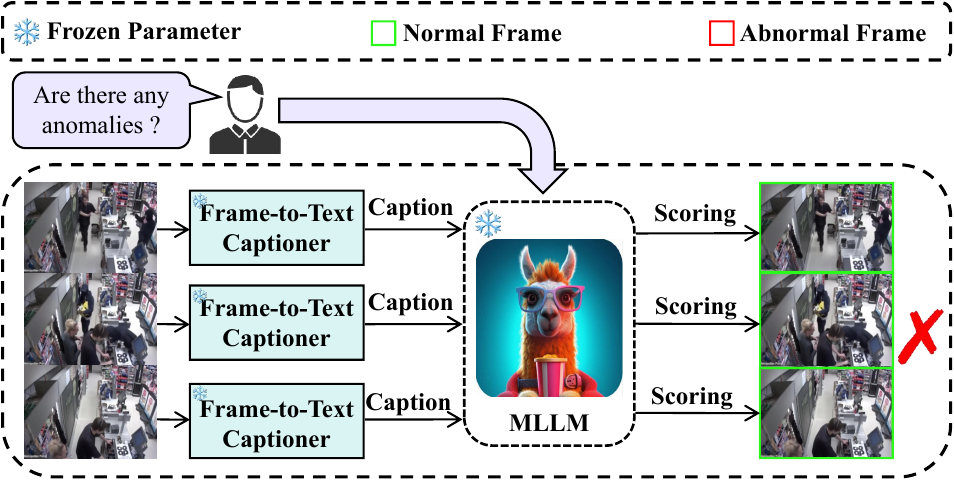}  % 调大宽度至通栏
  \label{fig:subfig1.1}
  }\\ % 添加换行符实现垂直排列
  
  \subfloat[Event-Aware Training-Free Video Anomaly Detection(ours)]{
  \includegraphics[width=0.45\textwidth]{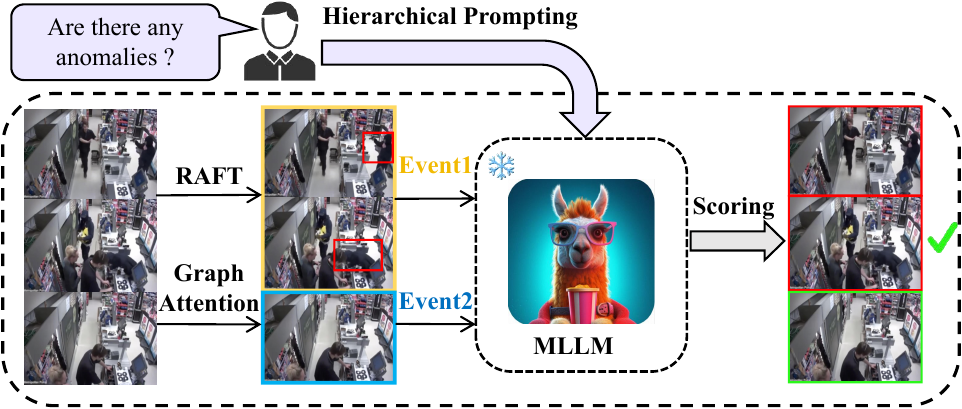}  % 调大宽度至通栏
  \label{fig:subfig1.2}
  }
  
  % \vspace{-0.3cm}
  \caption{Existing methods struggle to localize fine-grained abnormal frames. EventVAD segments events to enhance event consistency and fine-grained abnormal detection.}
  % \vspace{-0.5cm}
\end{figure}

\section{Introduction}
Recent years have witnessed a growing trend in employing large language models (LLMs) for multi-modal understanding across various multimedia applications. ~\cite{liu2023visual,zeng2025parameter,yan20243dsceneeditor,long2024generative,long2024trust}.
Among these applications, video anomaly detection (VAD)~\cite{ramachandra2020survey} has emerged as an important research direction to automatically identify and localize anomalous events or segments within video sequences~\cite{zeng2025local, zeng2025towards}.
% Leveraging multimodal large models (MLLMs)~\cite{liu2023visual,zeng2025parameter} to comprehend details in videos is crucial for multimedia \cite{yan20243dsceneeditor}. 
% Among them, Video Anomaly Detection~(VAD)~\cite{ramachandra2020survey} aims to locate the anomaly frames accurately ~\cite{zeng2025local, zeng2025towards}. 
Before the advent of LLMs~\cite{touvron2023llama,grattafiori2024llama,liu2023improved,zeng2024modalprompt,cheng2024videollama}, supervised training-based methods~\cite{sultani2018real,zaheer2020claws,wu2020not,wu2021learning,shao2024accidentblip} employ neural networks to achieve high precision (exceeding 90\%) on specific datasets.
%use neural networks to achieve greater precision 90\% on specific datasets.
However, these methods struggle to generalize in unseen settings~\cite{shao2025context} and typically require in-domain fine-tuning when applied to new scenarios.
Weakly-supervised\cite{sultani2018real,wang2019gods} or unsupervised methods\cite{zaheer2022generative} utilize single-class learning or auto-encoding strategies to train their models, but cannot achieve comparable results due to the shortage of direct and high-quality supervision signals.
% due to a shortage of labeled data, they cannot achieve performance comparable to that of supervised methods.

With the advancement of MLLMs~\cite{vaswani2017attention, zeng2024m2m,liu2023improved,zeng2024modalprompt, cheng2024videollama}, recent approaches have employed image-based models (e.g., LLaVA) to detect anomalous video segments via frame-by-frame analysis. 
Among these, LAVAD~\cite{zanella2024harnessing} represents the first attempt to propose a training-free framework, leveraging MLLMs to score anomalous frames without requiring any task-specific training.
Despite their success, existing approaches still face two key challenges:
(1) Image-based MLLMs often struggle to model temporal dynamics and diverse event patterns in videos, resulting in prediction inconsistency across frames.
(2) These methods (e.g., LAVAD) rely on a redundant multi-stage VQA pipeline~\cite{li2023blip,grattafiori2024llama,long2025retrieval}, which is inefficient in both storage and inference
~\cite{shao2024gwq, zeng2025mambaic, shao2025tr}.
% recent methods utilize
% image-based models (e.g., LLAVA)~\cite{vaswani2017attention, zeng2024m2m,liu2023improved,zeng2024modalprompt, cheng2024videollama} to detect anomalous video segments through frame-by-frame processing.
% However, these methods often fail to capture temporal transitions and diverse event patterns in videos, leading to inconsistent predictions.
% % struggle to understand fine-grained temporal content in video, which results in a lack of consistent understanding of videos. 
% As for training-free methods, LAVAD~\cite{zanella2024harnessing} leverages a combination of visual question-answering models~\cite{li2023blip} and LLMs~\cite{grattafiori2024llama} to score anomalous frames, thus achieving the localization of such frames. 
% LAVAD utilizes four image question-answering models to analyze each frame while relying on an LLM with at least 13 billion parameters for anomaly scoring. 
% This leads to an inefficient framework~\cite{shao2024gwq, zeng2025mambaic, shao2025tr} and may lead to long-tail problems.

Inspired by LAVAD~\cite{zanella2024harnessing}, we propose a training-free framework \textbf{EventVAD}, which segments long videos into short videos to enhance event consistency during MLLM understanding. 
EventVAD first extracts features from each frame using the ViT-B/16 of EVA-CLIP~\cite{sun2023eva}, and then augments inter-frame features using RAFT optical flow~\cite{teed2020raft} to capture fine-grained features. 
For temporal features, EventVAD employs the Graph Attention mechanism to aggregate and refine temporal features across frames. 
% This mechanism enhances the temporal consistency of events, making it more conducive to boundary detection and event segmentation. 
This design improves event consistency over time, helping detect boundaries and segment events more accurately.
Finally, EventVAD inputs the segmented events into an MLLM~\cite{cheng2024videollama} for scoring, to determine the anomaly level of each frame.

We evaluate EventVAD on UCF-Crime~\cite{sultani2018real} and XD-Violence~\cite{wu2020not} datasets. Experimental results demonstrate that EventVAD achieves \textit{state-of-the-art (SOTA)} compared to training-free, self-supervised, and one-class supervised methods, and even outperforms some weak-supervised methods. 
Moreover, our 7B-parameter model is smaller than 13B-parameter baselines (e.g., LAVAD), leading to more efficient storage and lower inference costs.
% we have reduced the model parameters from over 13 billion to 7 billion, significantly reducing storage and inference costs. 

Our contributions can be summarized as follows:
\begin{itemize}
    \item We propose a training-free event-aware video anomaly detection framework, \textbf{EventVAD}, which reduces errors in MLLMs scoring by segmenting events into short events to enhance videos' temporal consistency.
    \item We design a Graph Attention mechanism to model features across multiple frames. By incorporating RAFT optical flow to enhance frame features, Graph Attention enables more accurate event boundary detection and event segmentation.
    \item EventVAD achieves \textit{SOTA} performance among training-free, unsupervised, and one-class methods on both UCF-Crime and XD-Violence datasets, and even surpasses some weakly supervised methods. Additionally, EventVAD compresses the model's parameters from 13 to 7 billion, significantly reducing computational costs.
\end{itemize}

\section{Related Work}
\subsection{Video Anomaly Detection}
Video Anomaly Detection~(VAD) has gained significant attention in recent years due to its wide range of applications aimed at identifying anomalous events in video sequences~\cite{sultani2018real, nayak2021comprehensive, ramachandra2020survey, wang2025unifying, liao2025gm}.  Most research in VAD has focused on weakly supervised training paradigms, unsupervised training paradigms and training-free paradigms. Weakly-supervised VAD leverages video-level labels to localize frame-level anomalies, avoiding the prohibitive cost of manual frame annotation~\cite{joo2023clip, li2022scale, li2022self, sultani2018real, tian2021weakly, wu2021learning}. Most of these methods utilize 3D convolutional neural networks for feature learning and are trained with multi-instance learning~(MIL) loss \cite{yan2024renderworld,remondino2023critical,yan2025learning, yan2023nerfbk}. Unsupervised VAD methods do not require any labeled data and instead rely on the inherent structure of normal video data to detect anomalies~\cite{thakare2023dyannet, thakare2023rareanom, tur2023exploring, tur2023unsupervised, zaheer2022generative}. Among them, using generative models to capture normal data patterns in videos performs best in unsupervised video anomaly detection, especially in complex scenes and time series data processing. However, these training-based approaches require data collection and model retraining in practical deployment scenarios. As the first language-based method for training-free VAD using LLMs, LAVAD~\cite{zanella2024harnessing} leverages modality-aligned MLLMs to query and enhance the anomaly scores generated by large language models~(LLMs). Based on this, MCANet~\cite{dev2025mcanet} is a training-free video anomaly detection framework that dynamically generates multimodal text descriptions and analyzes them for efficient anomaly detection by fusing visual language models, audio language models and large language models. Building upon these innovations, through temporal segmentation into discrete event units, EventVAD significantly mitigates error propagation in long-form video analysis, reducing false detection rates.

\subsection{Video Understanding}
Video understanding tasks have been driven by the integration of MLLMs with temporal modeling. Early approaches in Video Large Language Models~(Video-LLMs)~\cite{zhang2023video, maaz2023video, lin2023video, li2024mvbench, ataallah2024minigpt4, jin2024video, liu2024st} primarily adopted vision-language adapters from Image-LLMs, such as cross-attention~\cite{long2021position,alayrac2022flamingo}, Q-former~\cite{li2023blip, zhu2023languagebind, ye2024mplug}, linear projection~\cite{liu2023visual, chen2023shikra, wang2024cogvlm} and dynamic visual tokenizer~\cite{jin2023unified}. These adapters bridge visual features with language models by compressing frame-level representations into fixed-length tokens. For instance, Video-LLaMA~\cite{zhang2023video} employs a video-specific Q-Former to aggregate temporal features, while Video-ChatGPT~\cite{maaz2023video} incorporates spatial and temporal pooling layers to enhance frame-level reasoning. Recent works like VTimeLLM~\cite{huang2024vtimellm} and CogVLM2~\cite{hong2024cogvlm2} propose boundary-aware training and timestamp-aware encoding, respectively, to address the problem of the limited effective context length of LLM for long video processing and redundancy in processing similar frames. The integration of multimodal alignment techniques has further expanded video understanding capabilities. Models like ImageBind~\cite{girdhar2023imagebind} unify embeddings across six modalities into a shared latent space, enabling zero-shot cross-modal retrieval. This approach has been widely adopted in frameworks such as PandaGPT~\cite{su2023pandagpt}, where audio streams are aligned with visual features via pre-trained encoders. However, these methods prioritize modality fusion over temporal-audio synchronization, limiting their ability to correlate dynamic events with acoustic cues. EventVAD leverages video LLMs to analyze segmented events, generating frame-level anomaly scores for precise temporal localization of abnormal content.

\begin{figure*}[t]
    \centering
    \includegraphics[width=0.9\linewidth]{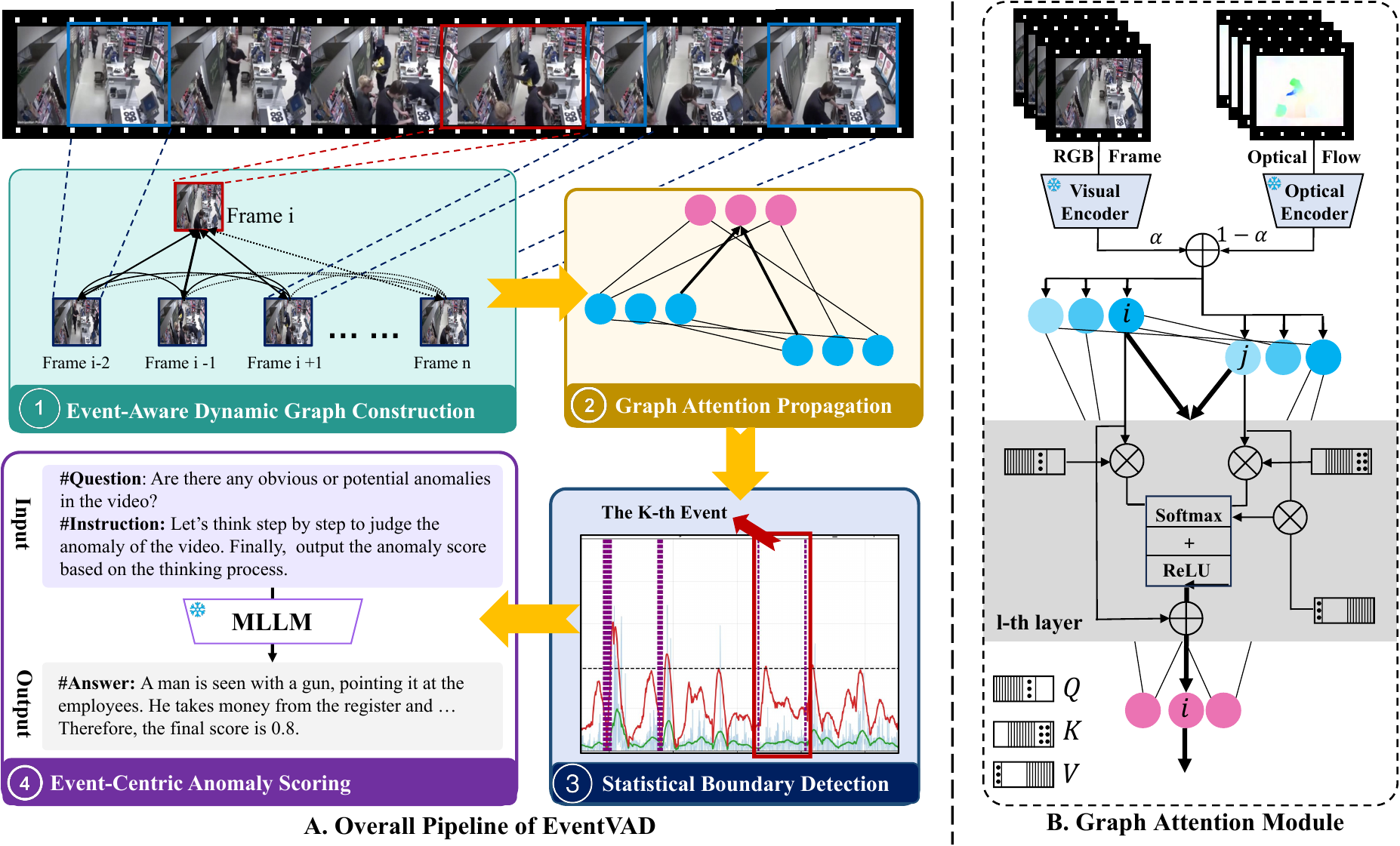}
    \caption{EventVAD comprises four core modules: (1) Dynamic Graph Construction for establishing spatiotemporal correlations, (2) Graph Attention Propagation for refining node features, (3) Statistical Boundary Detection for identifying event transitions and boundaries, and (4) Event-Centric Scoring for evaluating anomalies via MLLMs within semantic event units.
    }
    % \caption{Overall Pipeline of EventVAD. The Dynamic Graph Construction stage establishes spatiotemporal correlations using multimodal node features and time-decayed edges. Next, the Graph Attention Propagation refines node representations via orthogonally constrained message passing to enhance event transitions. Subsequently, the Statistical Boundary Detection stage identifies event boundaries by combining composite divergence metrics with adaptive thresholding. Finally, the Event-Centric Scoring stage employs MLLMs to evaluate anomaly likelihoods within semantically coherent event units.}
    \label{fig:pipeline}
    % \vspace{-0.2cm}
\end{figure*}

\section{Methodology}

% To address the neglect of event-centric semantic consistency in existing video anomaly detection methods, we propose an event-aware framework for anomaly frame detection.
Given an input video sequence $\mathcal{V} = \{\mathbf{I}_t\}_{t=1}^T$ with $T$ frames, our objective is to detect anomaly frames. $\{a_t\}_{t=1}^T$ are anomaly indicators, where $a_t \in \{0,1\}$ represents the binary label for frame $\mathbf{I}_t$. The overall pipeline is shown as Fig~\ref{fig:pipeline}, which consists of four components. Specifically, our method first constructs a dynamic graph with multimodal temporal decay, where nodes represent fused frame features and edges denote spatiotemporal connections. Then, graph attention propagation refines node features through orthogonally constrained messaging. Finally, event-centric scoring uses semantic cues to assess anomalies within detected event units.

\subsection{Event-Aware Dynamic Graph Construction}
\label{sec:Event-Aware Dynamic Graph Constructio}
\textbf{Node Feature Construction.}
To address the inherent limitations of static graphs in capturing dynamic video events and the slow training speed of graph neural networks ~\cite{scarselli2008graph}, we propose a dynamic spatiotemporal graph model based on video frames. Our model jointly optimizes multimodal node representations and time-sensitive edge connections. The proposed dynamic spatiotemporal graph model $G=(F,~E)$ introduces adaptive node representations through multimodal feature interaction. The vertex set F corresponds to video frame units with enhanced feature encoding, while edges in $E$ are dynamically established using cross-modal similarity metrics.

The nodal feature representation integrates a dual-stream architecture that combines semantic and motion patterns. As shown in Eq.~\eqref{E1}, the semantic branch employs CLIP~\cite{radford2021learning} to generate discriminative embeddings via L2-normalization.

\begin{equation}
{f_{i}^{\text{clip}}} = \frac{\phi_{\text{CLIP}}(\mathbf{I}_i)}{\|\phi_{\text{CLIP}}(\mathbf{I}_i)\|_2},
\label{E1}
\end{equation}
where \(\phi_{\text{CLIP}}\) generates 512-dimensional semantic embeddings. This explicit normalization amplifies feature discontinuities between frames, a critical component for detecting semantic-level event boundaries.

The motion branch captures temporal dynamics using optical flow processing. This branch is formulated as Eq.~\eqref{E2}:

\begin{equation}
{f_{i}^{\text{flow}}} = \mathbf{P}^\top \mathbb{E}[\mathbf{o}_{i{\to}i-1}],
\label{E2}
\end{equation}
where \(\mathbf{o}_{i{\to}i-1} \in \mathbb{R}^{H\times W\times 2}\) represents the optical flow fields from RAFT~\cite{teed2020raft}. The projection matrix \(\mathbf{P} \in \mathbb{R}^{2\times128}\) maps high-dimensional statistical features to 128-dimensional compact vectors.
% This dimensionality reduction improves the discriminability of motion patterns while preserving temporal coherence.

The fused node representation for the graph vertex $v_i \in F$ is calculated using
Eq.~\eqref{eq:Feature fusion}:
\begin{equation}\label{eq:Feature fusion}
\mathbf{f}_i =\alpha{f_{i}^{\text {clip}}} \oplus (1-\alpha){f_{i}^{\text {flow}}},
\end{equation}
where the semantic-motion fusion coefficient $\alpha=0.75$ and the eigenvector $\mathbf{f}_i$ integrate semantic and motion constraints to precisely define event transformation boundaries.
% where the feature vector $\mathbf{f}_i$ integrates semantic and motion constraints to precisely delimit event transformation boundaries.
% This characterization forms the basis for constructing edges in graph $G$, where temporal connectivity is determined by feature similarity metrics applied to the multimodal node representations.

\noindent\textbf{Temporal Decay Adjacency Matrix.}
% Fundamentally, analyzing the temporal evolution of video content requires integrating spatio-temporal variations across components. 
To capture relationships between nodes, we construct a dynamic graph with time decay.
% Specifically, prominent transition points—characterized by abrupt scene changes or significant shifts in motion patterns—define event patterns. 
The time decay adjacency matrix $E$ is defined as Eq.~\eqref{E4}:
\begin{equation}
E_{i j}=\frac{\alpha\cdot\cos \left(f_{i}^{\text {clip }}, f_{j}^{\text {clip }}\right)+(1-\alpha)\cdot\exp \left(-\| f_{i}^{\text {flow }}-f_{j}^{\text {flow }} \mid\right)}{1+\gamma \cdot|i-j|}.
\label{E4}
\end{equation}
where the semantic-motion fusion coefficient $\alpha=0.75$ is shared with the node usage coefficient. To emphasize short-term correlations and suppress long-term ones, our model combines multimodal similarity between frames $i$ and $j$ in the numerator. In the denominator, we introduce a $\gamma$ time decay factor to penalize remote associations, which diminish with temporal distance.
% Real-world events exhibit short-term aggregation; thus, the numerator combines multimodal similarities between frames $i$ and $j$ to emphasize short-term correlations and suppress long-term ones. In the denominator, contextual associations decay with temporal distance. We introduce a $\gamma$ time decay factor to penalize remote associations. As $\Delta t$ increases, hyperbolic decay suppresses pseudo-correlations between non-causal frames.

\subsection{Graph Attention Networks Propagation}
\label{sec:Graph Attention Networks Propagation}
% \noindent\textbf{Graph Attention Networks Propagation}
To improve frame-by-frame representation while preserving temporal consistency, we propose a training-free attention mechanism based on orthogonal feature projection in spacetime. This mechanism amplifies segment contrast through graph-guided message passing, enhancing event boundary distinguishability. Our approach builds upon graph attention networks~\cite{velivckovic2018graph} with orthogonal constraints.

Given the propagated node features $\mathbf{F}^{(0)} = [\mathbf{f}_1, ..., \mathbf{f}_n]^\top \in \mathbb{R}^{n\times d}$ from Section \ref{sec:Event-Aware Dynamic Graph Constructio}, we project the features into orthogonal subspaces to prevent dimension collapse as Eq~\eqref{E5},
\begin{equation}
\left\{\begin{matrix} \mathbf{Q} = \text{QR}\big\{\mathcal{N}(0,1)^{d\times k}\big\}\\\mathbf{K} = \text{QR}\big\{\mathcal{N}(0,1)^{d\times k}\big\}\\\mathbf{V} = \text{QR}\big\{\mathcal{N}(0,1)^{d\times d}\big\}\\\end{matrix}\right.,
\label{E5}
\end{equation}
where \text{QR(·)} denotes orthonormal columns via QR decomposition, \(d=640\) is the fused feature dimension, and \(k=64\) is the projected dimension. These fixed orthogonal matrices maximize feature retention.

%where \text{QR(·)} denotes QR decomposition ensuring orthonormal colu-mns, \(d=640\) is the dimension of the fused feature, \(k=64\) is the projected dimension. These fixed orthogonal matrices ensure maximum feature retention during propagation.

For node $f_i$ the attention weight for neighbor $f_j$ in the dynamic graph is calculated as:
\begin{equation}
Atten_{ij} = \text{Softmax}\left( \frac{ (\mathbf{f}_i\mathbf{Q})(\mathbf{f}_j\mathbf{K})^\top }{\sqrt{d_a}}\right)(\mathbf{f}_j^{(t)}\mathbf{V}).
\label{E3}
\end{equation}

The indicator function $\mathbb{E}_{(i,j)}$ corresponds to temporal connectivity in Section 3.1, ensuring attention respects event-induced topology. This constraint prevents attention dispersion to irrelevant frames. Features update iteratively via:
\begin{equation}
\mathbf{f}_i^{(t+1)} = \mathbf{f}_i^{(t)} + \sum_{j\in\mathcal{N}_i} Atten_{ij} \cdot \mathbb{E}_{(i,j)}.
\label{E7}
\end{equation}
 
After each iteration, features are centered as:
\begin{equation}
\mathbf{f}_i^{(t+1)} \leftarrow \mathbf{f}_i^{(t+1)} - \frac{1}{|F|}\sum_{k\in F} \mathbf{f}_k^{(t+1)}.
\label{E8}
\end{equation}

Through iterative graph attention updates, feature propagation maintains global divergence and local consistency, preserving relative differences critical for anomaly detection.

\subsection{Statistical Boundary Detection}
\label{sec:Statistical Boundary Detection}
For enhanced graph attention network features, we propose a statistical boundary detection mechanism to identify event transitions via temporal discontinuity analysis.

For consecutive frames $(i, i+1)$, we compute a composite divergence metric:
\begin{equation}
s_i = \underbrace{\|\mathbf{f}_{i+1} - \mathbf{f}_i\|_2^2}_{\text{Feature Magnitude Jump}} +\underbrace{\left(1 - \frac{\mathbf{f}_i^\top \mathbf{f}_{i+1}}{\|\mathbf{f}_i\| \|\mathbf{f}_{i+1}\|}\right)}_{\text{Cosine Dissimilarity}},
\label{E9}
\end{equation}
where the squared L2 term amplifies abrupt feature-space jumps, and the cosine term detects directional changes in the manifold. This dual strategy captures amplitude and directional discontinuities caused by anomalies.

The raw dissimilarity score $s_i$ includes high-frequency noise from camera shake or minor motion. To suppress noise while preserving true boundaries, we apply a Savitzky-Golay filter with a window width $w=60$, defined as:
\begin{equation}
\tilde{s}_i = \frac{1}{w}\sum_{k=-w/2}^{w/2} a_k s_{i+k},
\label{E10}
\end{equation}
where $a_k$ derives from a quadratic polynomial fit, retaining true discontinuities.

For the smoothed signal $\tilde{s}_i$, we compute a moving average:
\begin{equation}
\mu_i = \frac{1}{w} \sum_{k=i-\lfloor w/2 \rfloor}^{i+\lfloor w/2 \rfloor} \tilde{s}_i.
\label{Mui}
\end{equation}
The signal ratio $r_{i}$ is then calculated as:
\begin{equation}
r_{i} =\frac{\tilde{s}_i}{\mu_i},
\label{E11}
\end{equation}
where equal window weights suppress transient fluctuations, enhancing noise robustness.

To mitigate outlier sensitivity from occlusions or jitter, we use median absolute deviation for adaptive thresholding:
\begin{equation}
\operatorname{MAD}(r_{i})=\operatorname{median}(|r_{i}-\operatorname{median}(r_{i})|),
\label{E12}
\end{equation}
\begin{equation}
\operatorname{M}=\operatorname{median}(r_{i})+k\times\operatorname{MAD}(r_{i}).
\label{E13}
\end{equation}
Adjusting $k=3$ aligns with the $3\sigma$ principle~\cite{leys2013detecting}, capturing 99.7\% of normal variation. Boundaries are identified when 
$r_{i}>M$.
Finally, adjacent boundary points are merged to prevent over-segmentation, refining event boundary accuracy.

\subsection{Event-Centric Anomaly Scoring}
\label{sec:svent-Centric Anomaly Scoring}
% Previous video anomaly detection methods~\cite{zanella2024harnessing} often use discrete frame-level analysis or global video processing strategies, but due to the non-adaptive segmentation mechanism, too short time fragments will lead to the loss of temporal information, and too long video units will weaken the spatial feature representation, which restricts the visual language model's ability to understand in video.
Existing video anomaly detection methods~\cite{zanella2024harnessing} often rely on discrete frame-level analysis or global video processing. However, their non-adaptive segmentation mechanisms face a critical trade-off: overly short temporal fragments lose contextual information, while excessively long units weaken spatial feature representation. This limitation hinders visual language models' ability to interpret videos effectively.

To address this, we construct event semantic units as spatiotemporal primitives, creating an optimized feature representation framework for visual-language models~\cite{liu2023visual}. For event unit analysis, we propose a semantic-driven hierarchical prompting framework. This framework directs multimodal large language models to produce structured outputs: first generating video content descriptions and then deriving anomaly scores. When processing a video, the multimodal large language models initially generate descriptive text by identifying surface and latent semantic features. It subsequently outputs an anomaly score based on this description, enabling cross-modal evaluation against predefined criteria.

By implementing a two-stage reasoning framework, we establish a self-correction mechanism. This architecture allows systematic score derivation through video content analysis, ensuring contextual relevance and reducing scoring inaccuracies.
%\begin{figure*}[htbp]
%  \centering
%  \includegraphics[width=0.9\textwidth]{3.4 picture.png}
%  \caption{An example of a visual two-stage inference mechanism, the first stage generates descriptive text based on visual semantics, and the second stage realizes the synchronous generation of anomaly scores based on descriptive text. There is a red bounding box in the image that highlights key image areas to provide the necessary information to answer the question.}
%  \label{fig:model_score}
%\end{figure*}

\section{Experiment}
% We evaluated EventVAD on two benchmarks. 
% We introduce the experimental setups in Sec.~\ref{sec:setup}, demonstrate main results in Sec.~\ref{sec:main-results}, and conduct ablation studies in Sec.~\ref{sec:ablation-study}.
% Sec.~\ref{sec:setup}, we detail the experimental setup in terms of data sets and performance metrics, then present and discuss the results in Section~\ref{sec:main-results}, and finally, the ablation experiments in Section~\ref{sec:ablation-study}.

\subsection{Experiment Setting}
\label{sec:setup}
\noindent \textbf{Setting Up.}
% Our foundation model employs Video-LLaMA2~\cite{cheng2024videollama}, enabling precise video comprehension. All experiments were conducted on single NVIDIA A800 (80GB) GPU.
We adopt the CLIP ViT-B/16 model~\cite{radford2021learning} to generate 512-dimensional semantic embeddings, and the pre-trained RAFT model to obtain 128-dimensional optical flow features, with $FPS=30$.
We also use bilinear interpolation for spatial alignment to maintain resolution consistency. 
% The encoder uses a dual-branch architecture that combines visual semantics and motion representation, using the CLIP ViT-B/16 model~\cite{radford2021learning} to generate 512-dimensional semantic embeddings, while optical flow features are extracted by the pre-trained RAFT model to generate 128-dimensional vectors, with frame pairs processed at intervals Δt=1, and bilinear interpolation for spatial alignment to maintain resolution consistency. 
% The scoring architecture was implemented using 
We employ the VideoLLaMA2.1-7B-16F model ~\cite{cheng2024videollama} as the backbone model.
Additionally, the time decay factor is set to $\gamma=0.6$, the semantic-motion fusion coefficient is set to $\alpha=0.75$, and the graph attention propagation is only a single iteration. All experiments were conducted on a single NVIDIA A800 (80GB) GPU.

\noindent \textbf{Datasets.} 
We evaluate EventVAD on two widely adopted video anomaly detection benchmarks, UCF-Crime~\cite{sultani2018real} and XD-Violence~\cite{wu2020not}. 
% The details of datasets are as Appendix~\ref{data}.

\noindent \textbf{Evaluation Metrics.}
For the UCF-Crime dataset, we employ frame-level \textbf{AUC-ROC} (Area Under the Receiver Operating Characteristic Curve). For the XD-Violence dataset, we also employ \textbf{AP} (Average Precision) as the primary metric to align with standard VAD settings.
% The detail can be seen in Appendix~\ref{Evaluation Metrics}. 

\subsection{Main Results}
\label{sec:main-results}
\textbf{Results on UCF-Crime.}
Since the UCF-Crime data comes from the real shooting scene of the camera, the similarity between events in the same video is high. It is more challenging when the model divides the events. As shown in Tab.~\ref{table:ucf_results}, \textbf{EventVAD surpasses all training-free methods to \textit{state-of-the-art (SOTA)} and outperforms LAVAD by nearly 4\%.} compared to LAVAD. EventVAD's foundational model is Video-Llama2~\cite{cheng2024videollama} with a parameter count of 7b, which has a significant decrease in the overall parameter of the model. Also, EventVAD outperforms all unsupervised and one-class methods, as well as some weakly supervised methods on UCF-Crime. 

By analyzing the results, we found that since the other methods ignore dividing the video, different model parameter sizes cannot achieve accurate localization when dealing with long videos. This suggests that EventVAD's approach of dividing events and then scoring event anomalies can effectively mitigate the model's inability to achieve anomaly detection for long videos. Even in noisy scenarios, EventVAD demonstrates robust detection capability under challenging conditions with both rare abnormal samples (low occurrence rate <5\%) and significant visual noise.

\begin{table}[t!]
\resizebox{1\linewidth}{!}{
\centering
\begin{tabular}{lccc}
\toprule
\textbf{Method} & \textbf{Backbone} &\textbf{Supervised}& \textbf{AUC (\%)} \\
\midrule
Wu \textit{et al.}~\cite{wu2020not} & I3D-RGB&Weakly Supervised & 82.44\\
RTFM~\cite{tian2021weakly} & I3D-RGB &Weakly Supervised& 84.03\\
Wu $\&$ Liu \textit{et al.}~\cite{wu2021learning} & I3D-RGB &Weakly Supervised& 84.89\\
MSL~\cite{li2022self} & VideoSwin-RGB &Weakly Supervised& 85.62\\
S3R~\cite{wu2022self} & I3D-RGB &Weakly Supervised& 85.99\\
MGFN~\cite{chen2023mgfn} & I3D-RGB&Weakly Supervised & 86.98\\
SSRL~\cite{li2022scale} & I3D-RGB &Weakly Supervised& 87.43\\
CLIP-TSA~\cite{joo2023clip} & ViT&Weakly Supervised & 87.58 \\
\midrule
SVM~\cite{sultani2018real} & -& One Class& 50.00\\
SSV~\cite{sohrab2018subspace} & - &One Class& 58.50\\
BODS~\cite{wang2019gods} & I3D-RGB & One Class&68.26\\
GODS~\cite{wang2019gods} & I3D-RGB &One Class& 70.46\\
\midrule
GCL~\cite{zaheer2022generative} & ResNext &Unsupervised& 74.20\\
Tur~\cite{tur2023exploring} & ResNet &Unsupervised & 65.22\\
Tur~\cite{tur2023unsupervised} & ResNet  &Unsupervised& 66.85\\
DyAnNet~\cite{thakare2023dyannet} & I3D  &Unsupervised& 79.76\\
\midrule
Blip2~\cite{li2023blip}&ViT&Training Free&46.42\\
ZS CLIP~\cite{radford2021learning} & ViT &Training Free & 53.16\\
ZS ImageBind (Image)~\cite{girdhar2023imagebind} & ViT&Training Free & 53.65\\
ZS ImageBind (Video)~\cite{girdhar2023imagebind} & ViT&Training Free & 55.78\\

LLaVA-1.5~\cite{liu2023improved} & ViT&Training Free & 72.84\\
Video-Llama2~\cite{zhang2023video}& ViT&Training Free & 74.42\\
LAVAD~\cite{zanella2024harnessing} &  ViT&Training Free & 78.33\\
\textbf{EventVAD (Ours)} &  \textbf{ViT}&\textbf{Training Free} & \textbf{82.03}\\
\bottomrule
\end{tabular}
}
\caption{Results on the UCF-Crime dataset.}
\label{table:ucf_results}
 \vspace{-1.0cm}
\end{table}

\noindent \textbf{Results on XD-Violence.}
Since the XD-Violence datasets are all derived from video recordings, the videos are clearer, and there is less noise in the picture. As shown in Tab.~\ref{table:xd_violence_results}, \textbf{EventVAD's AP and AUC on this dataset are almost 5\% higher than the current training-free SOTA LAVAD~\cite{zanella2024harnessing}.} Also, EventVAD anomaly detection capability surpasses all one-class and unsupervised methods. This indicates that EventVAD is not only suitable for low-resolution scenarios but also applicable to high-resolution scenarios. Even though EventVAD has significantly fewer model parameters than LAVAD, its performance shows a substantial improvement compared to LAVAD. This can significantly reduce computational and deployment costs for the model. 

After analyzing the samples that LAVAD failed to detect, we found that the current model performs poorly in long-video detection. Additionally, many abnormal events only occupy a small portion of the frame, making it difficult for the model to achieve precise localization. Moreover, compared to LAVAD, EventVAD's prompt setup is very straightforward. We achieve multi-stage reasoning in MLLM through hierarchical prompting to enhance its scene-understanding capability.

\begin{figure}[t]
  \centering
  \subfloat[Abnormal video before graph attention networks propagation]{\includegraphics[width=0.48\columnwidth]{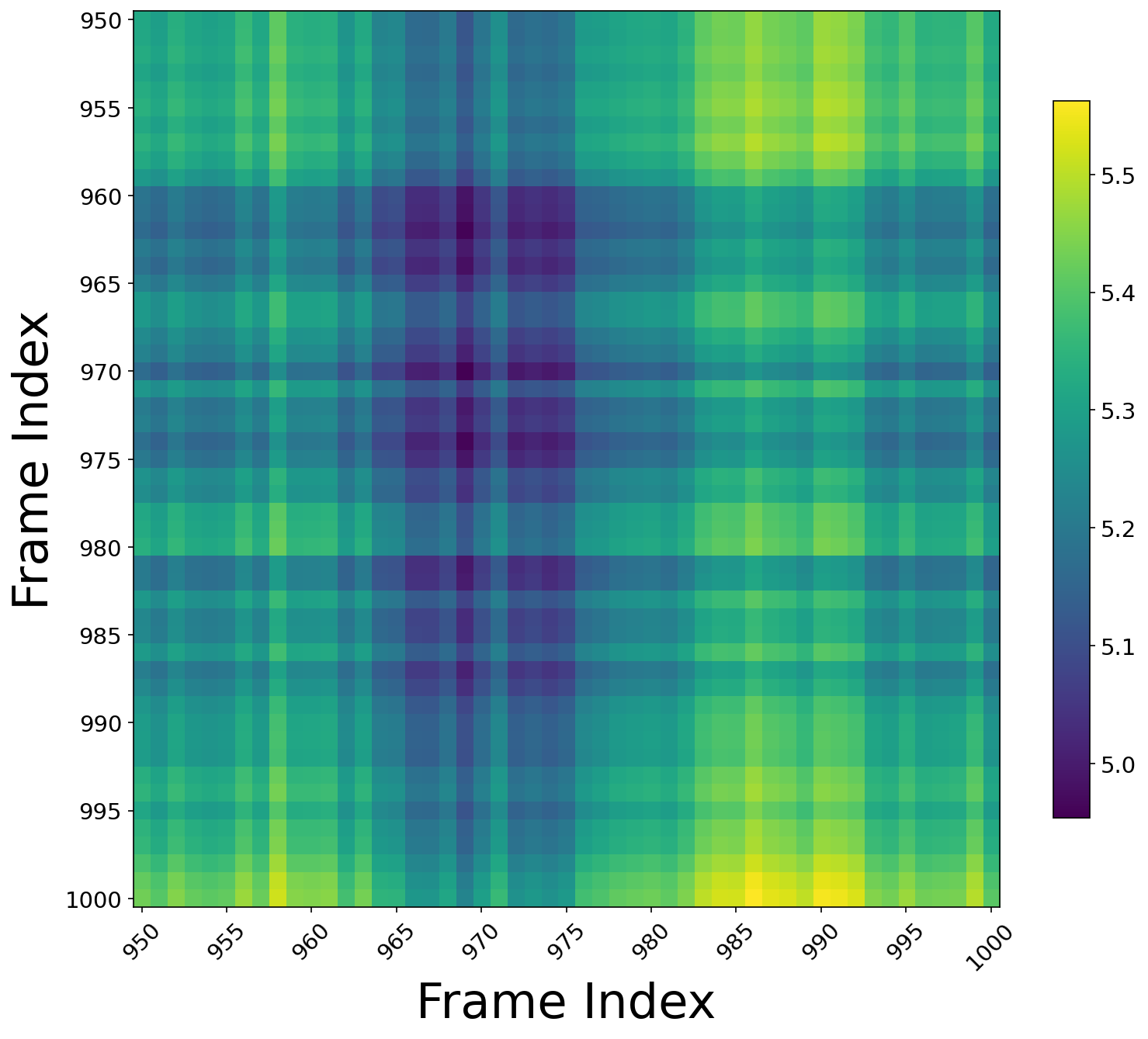}\label{fig:subfig11}}
  \hspace{0.02\columnwidth}
  \subfloat[Abnormal video after graph attention networks propagation]{\includegraphics[width=0.48\columnwidth]{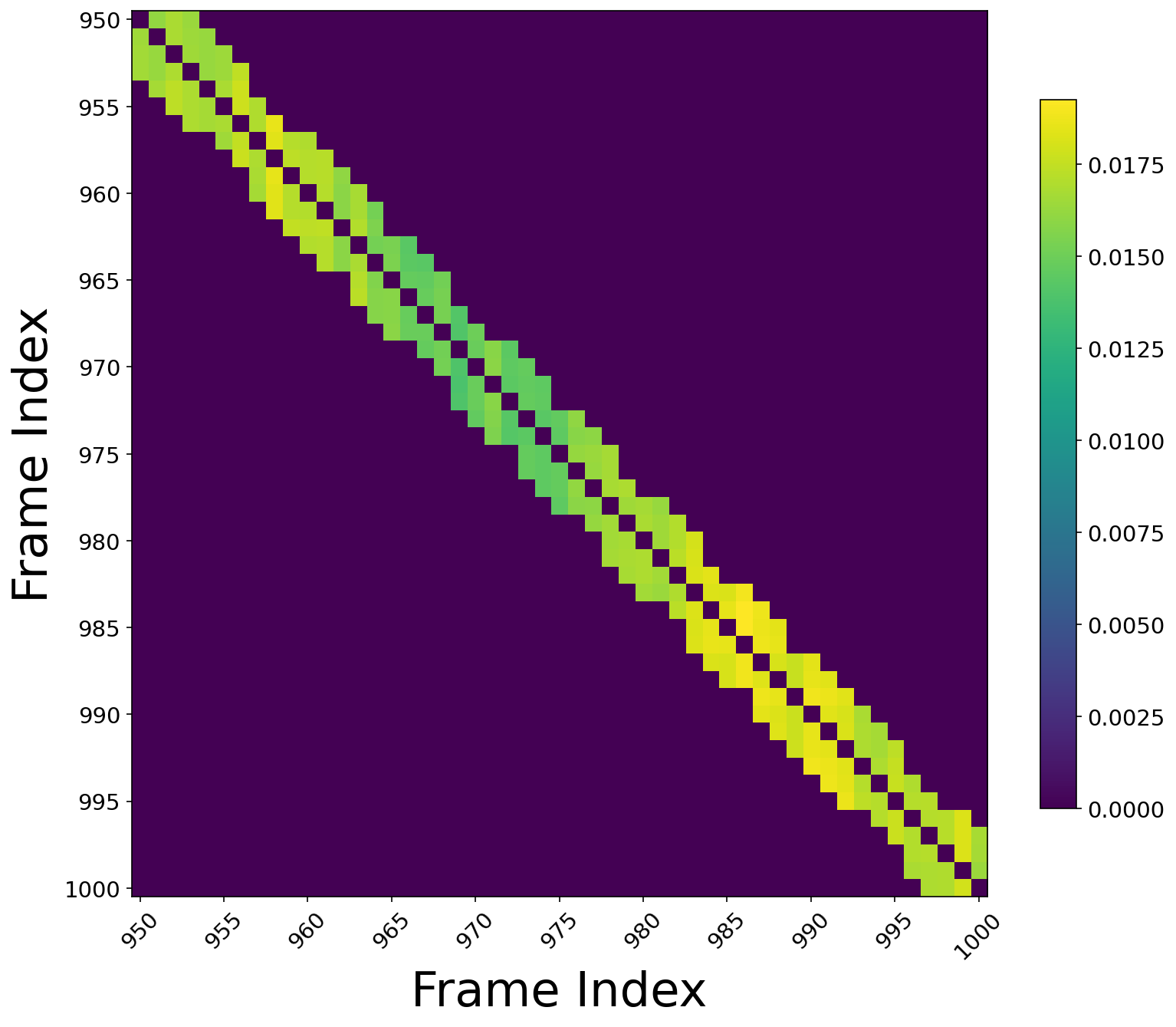}\label{fig:subfig12}} 
  
  % \vspace{0.5em}
  \subfloat[Normal video before graph attention networks propagation]{\includegraphics[width=0.48\columnwidth]{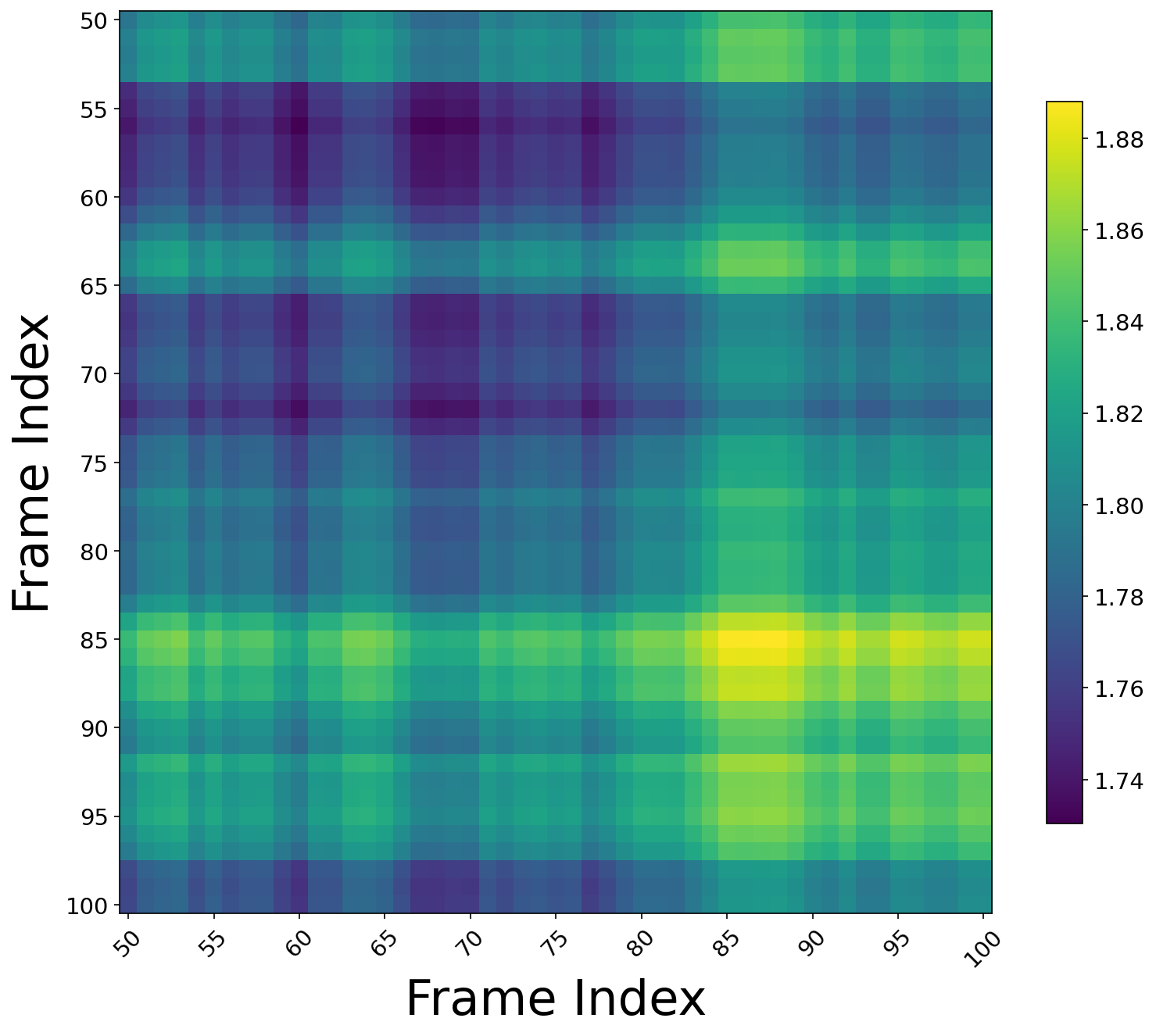}\label{fig:subfig13}}
  \hspace{0.02\columnwidth}
  \subfloat[Normal video after graph attention networks propagation]{\includegraphics[width=0.48\columnwidth]{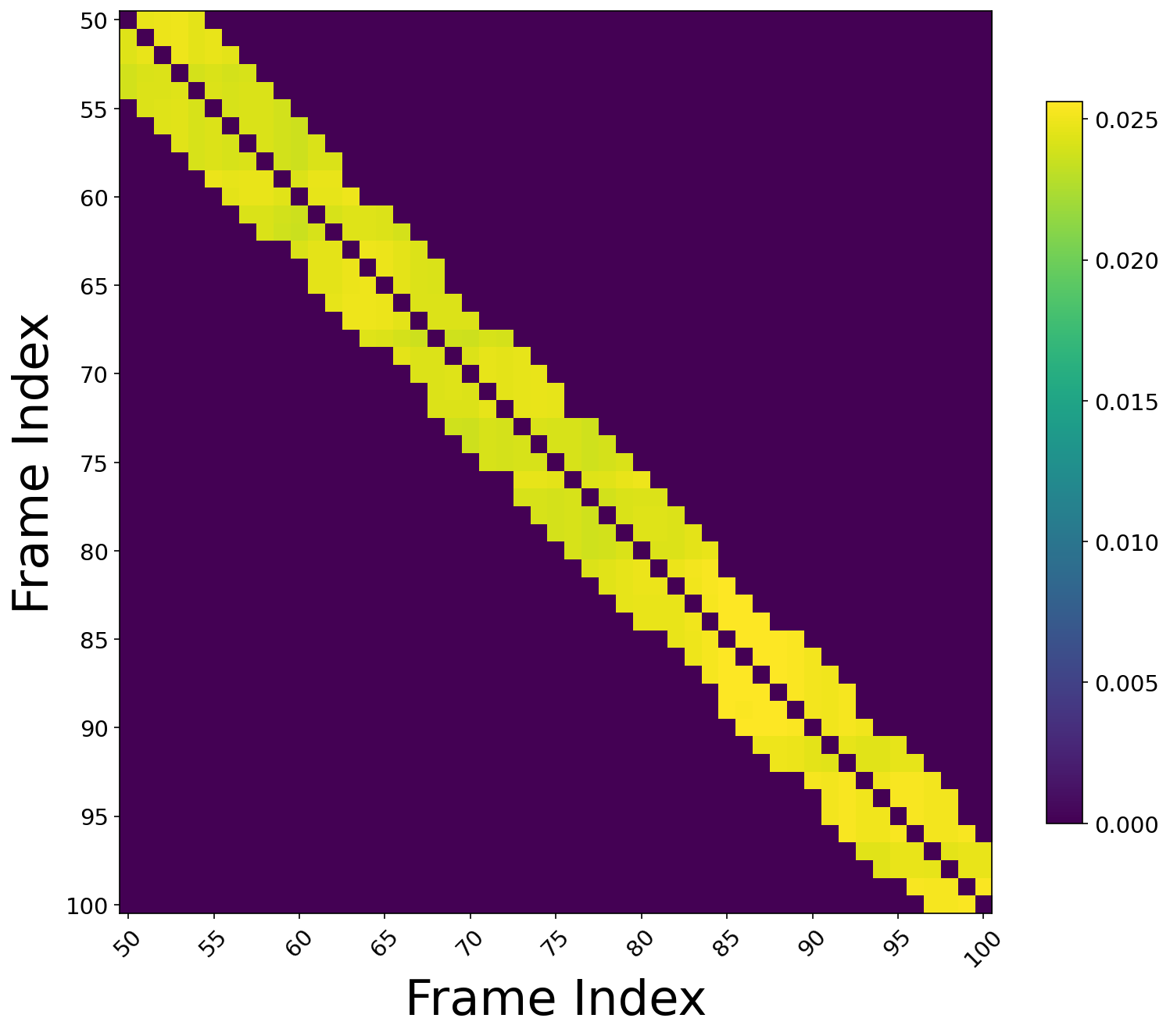}\label{fig:subfig14}}
  
  % \vspace{-0.3cm}
  \caption{Graph attention propagation visualization analysis. We take the abnormal and normal video samples in the UCF-Crime dataset to visualize the inter-frame relationship before and after the application of Graph Attention Propagation. The heat maps show the change of the weight relationship between frames in the corresponding frame interval.}
  % \vspace{-0.5cm}
  \label{fig:vis1}
\end{figure}

\begin{table}[t!]
\tabcolsep 3pt
\resizebox{\linewidth}{!}{
\centering
\begin{tabular}{lcccc}
\toprule
 \textbf{\textsc{Method}} & \textbf{ \textsc{Backbone}}&\textbf{Supervised} & \textbf{\textsc{AP (\%)}} & \textbf{\textsc{AUC (\%)}} \\
\midrule
Wu \textit{et al.}~\cite{wu2020not} & I3D-RGB &Weakly Supervised& 73.20 & -\\
Wu and Liu~\cite{wu2021learning} & I3D-RGB&Weakly Supervised & 75.90 & -\\
RTFM~\cite{tian2021weakly} & I3D-RGB&Weakly Supervised &77.81 & -\\
MSL~\cite{li2022self} & VideoSwin-RGB&Weakly Supervised & 78.58 & -\\
S3R~\cite{wu2022self} & I3D-RGB &Weakly Supervised& 80.26 & -\\
MGFN~\cite{chen2023mgfn} & VideoSwin-RGB &Weakly Supervised&80.11 & -\\
\midrule
Hasan \textit{et al.}~\cite{hasan2016learning} & \text{AE\textsuperscript{RGB}} & One Class& - & 50.32\\
Lu \textit{et al.}~\cite{lu2013abnormal} & Dictionary & One Class& - & 53.56\\
BODS~\cite{wang2019gods} & I3D-RGB & One Class& - & 57.32\\
GODS~\cite{wang2019gods} & I3D-RGB & One Class& - & 61.56\\
\midrule
RareAnom~\cite{thakare2023rareanom} & I3D-RGB &Unsupervised& - & 68.33\\
\midrule
Blip2~\cite{li2023blip}&ViT&Training Free&10.89&29.43\\
ZS CLIP~\cite{radford2021learning}  & ViT &Training Free& 17.83 & 38.21\\
ZS ImageBind (Image)~\cite{girdhar2023imagebind}  & ViT &Training Free& 27.25 & 58.81\\
ZS ImageBind (Video)~\cite{girdhar2023imagebind}  & ViT&Training Free & 25.36 & 55.06\\
LLaVA-1.5~\cite{liu2023improved} & ViT &Training Free& 50.26 & 79.62\\
Video-Llama2~\cite{zhang2023video}& ViT&Training Free & 53.57 & 80.21\\
LAVAD~\cite{zanella2024harnessing}&   ViT&Training Free & 60.02 & 82.89\\
\textbf{EventVAD (Ours)} &   \textbf{ViT}&\textbf{Training Free} & \textbf{64.04} & \textbf{87.51}\\
\bottomrule
\end{tabular}
}
\caption{Results on the XD-Violence dataset.}
\label{table:xd_violence_results}
	 \vspace{-0.5cm}
\end{table}
\noindent \textbf{Visual Analysis.}
As shown in Fig.~\ref{fig:vis1}, using a UCF-Crime video sample as an example, we demonstrate a comprehensive visualization of abnormal and normal sequences by the graph attention propagation process.

Fig.~\ref{fig:vis1} reveals fundamental differences in the modeling of the temporal relationship between normal and abnormal sequences. Both the Subgraph \ref{fig:subfig11} and Subgraph \ref{fig:subfig13} show chaotic inter-frame connections, with low discriminative long-distance dependence. After the attention propagation of the applied graph, subgraphs (c) and (d) clearly show strengthened connections between adjacent frames, and the pseudo-correlation with the distant frame is suppressed.
In particular, Subgraph \ref{fig:subfig14} as a whole presents a relatively uniform inter-frame weight, but Subgraph \ref{fig:subfig13} has a unique cluster of low weights around frames 965-975, which corresponds to the actual boundary division.

This actually verifies how the propagation process of our graph attention acts as a feature contrast amplifier, with the help of which the Statistical Boundary Detection module can convert the amplified signal difference into an accurate event boundary.

\begin{figure}[t]
  \centering
  \subfloat[Abnormal video statistical boundary detection]{\includegraphics[width=1\columnwidth]{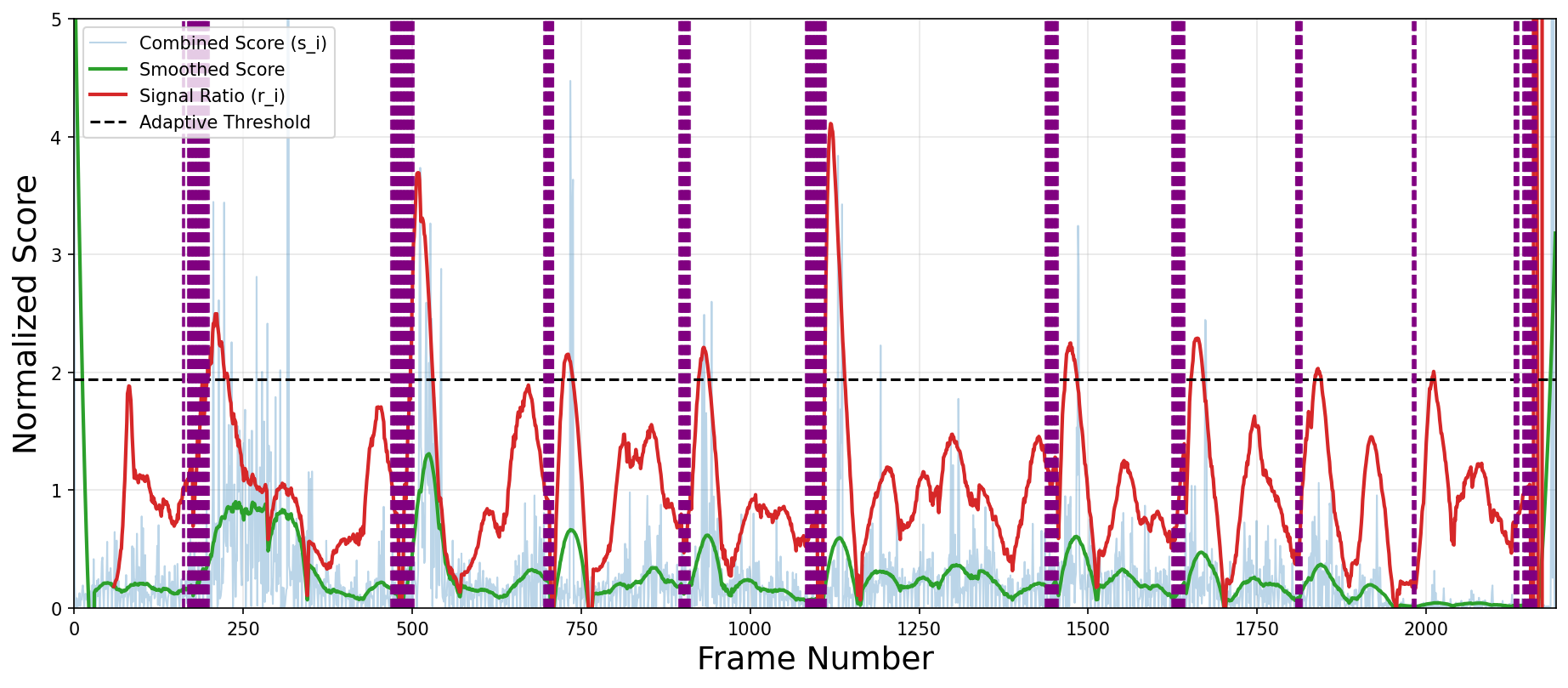}\label{fig:subfig5}}
  \hspace{0.02\columnwidth}
  
   \vspace{0.5em}
  \subfloat[Normal video statistical boundary detection]{\includegraphics[width=1\columnwidth]{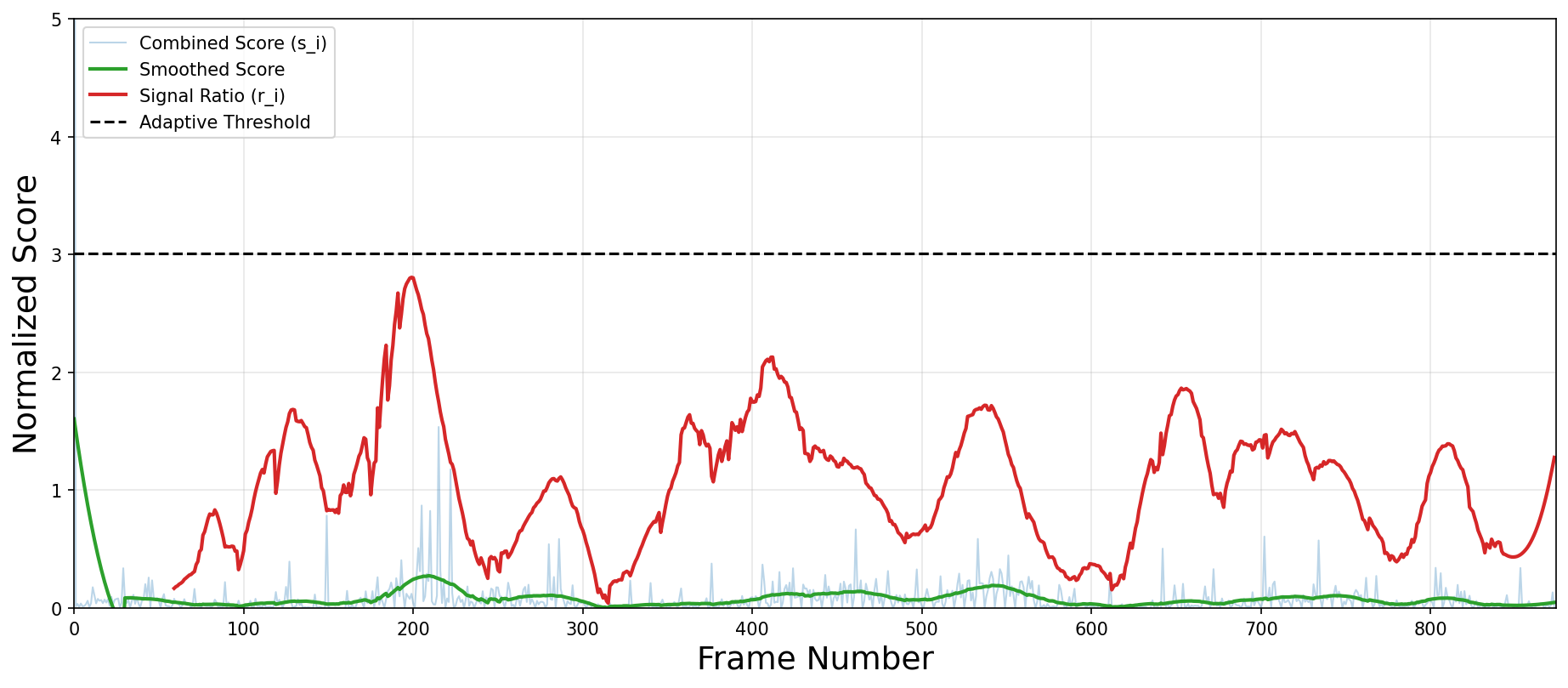}\label{fig:subfig6}}
  \hspace{0.02\columnwidth}
  
   \vspace{-0.3cm}
  \caption{Statistical boundary detection visual analysis. We use samples from the UCF-Crime dataset as an example to visualize the boundary detection process of abnormal and normal videos.}
   \vspace{-0.5cm}
  \label{fig:vis3}
\end{figure}

\noindent \textbf{Boundary Detection Visualization.}
As shown in Fig.~\ref{fig:vis3}, we visualize the boundary detection of all frames of these video samples. The light blue curve represents the original signal used for segmentation, the green curve represents the signal after being smoothed by Savitzky-Golay filtering, the red curve represents the change in the ratio of the smoothed signal within the window to the average line of the corresponding window, the black horizontal line represents the adaptive threshold, and the purple horizontal line represents the detected boundary.

For different types of videos, there are usually different event boundary patterns. Normal video sequences have fewer events due to their good consistency, while abnormal videos have the opposite. The difference between Subgraph \ref{fig:subfig5} and Subgraph \ref{fig:subfig6} validates that our framework can take advantage of this feature to obtain clear boundary points at different boundary densities when the score threshold is set to 2, thus achieving a robust distinction between normal time coherence and anomalous discontinuity.

In the Subgraph \ref{fig:subfig5}, for frames 965-975 of the video, our framework can effectively capture the key boundary detection patterns after the graph attention propagation, which also quantitatively verifies the subsequent boundary detection process of the Subgraph \ref{fig:subfig12}. In addition, the framework exhibits noise elasticity when dealing with high variance clips, such as 1200 to 1300 frames of video, and maintains some sensitivity to interframe variations in overall signal-stable clips, such as 1750 to 1900 frames of video.

\begin{figure*}[htbp]
  \centering
  \subfloat[Abnormal Examples in UCF-Crime Dataset]
  {\includegraphics[width=0.5\textwidth]{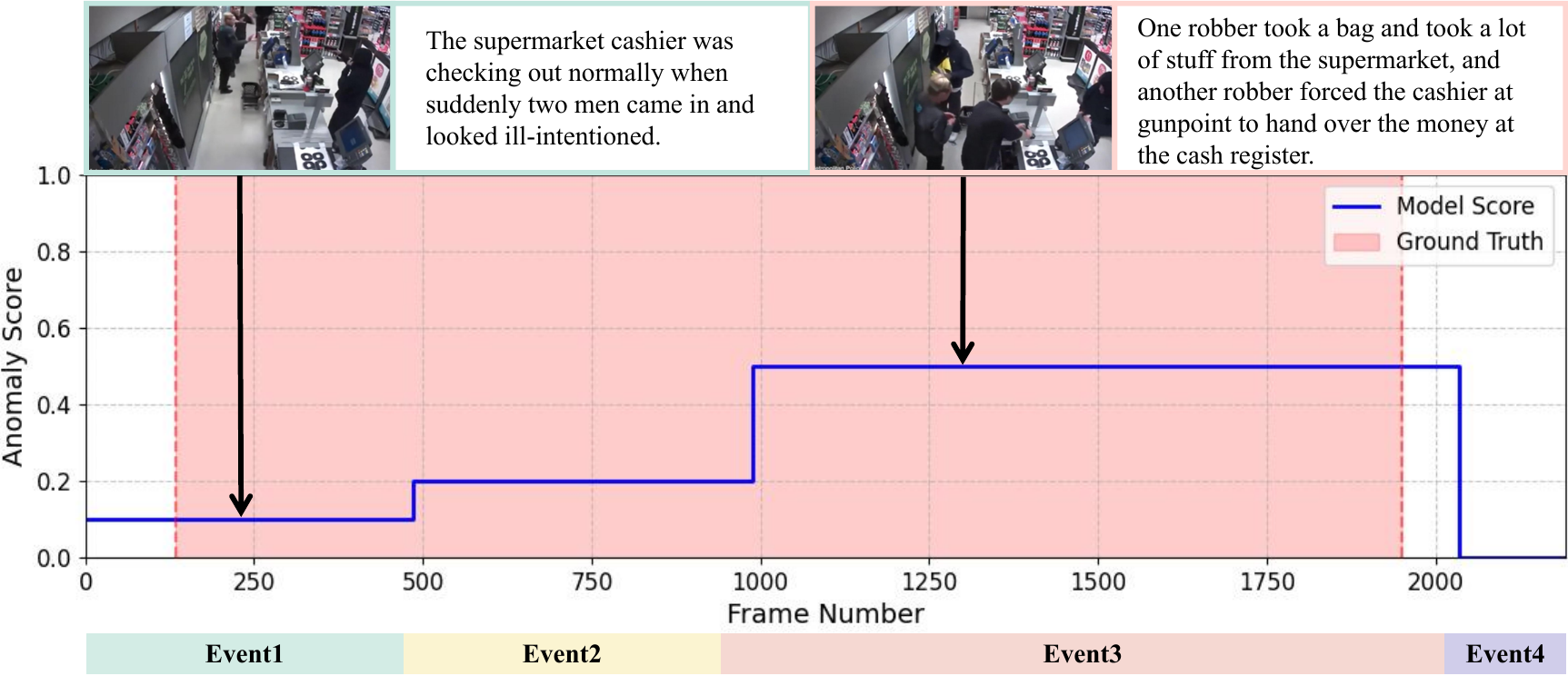}\label{fig:subfig7}}
  % 重点就在这，优先横向排列，自动换行
  \subfloat[Normal Examples in UCF-Crime Dataset]
  {\includegraphics[width=0.5\textwidth]{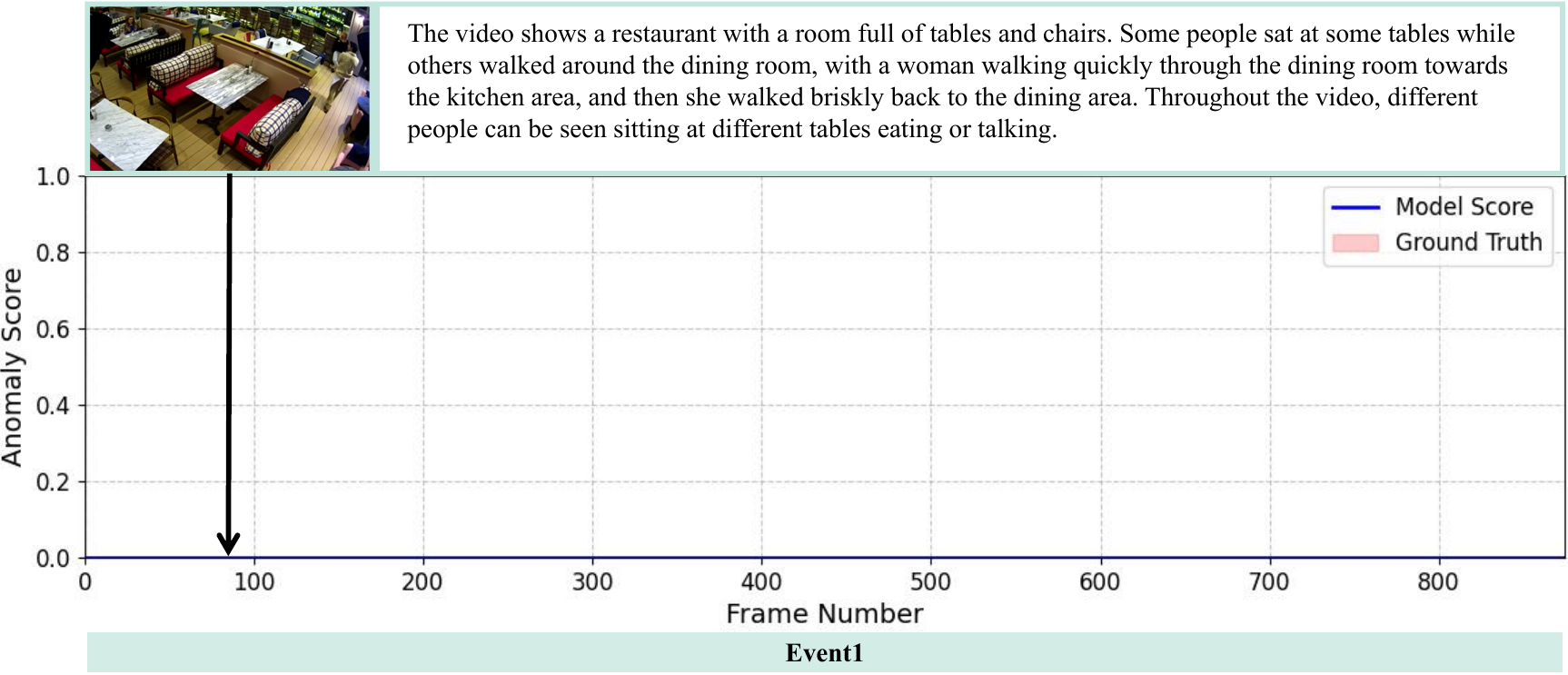}\label{fig:subfig8}}  
  
  \subfloat[Abnormal Examples in XD-Violence Dataset]
  {\includegraphics[width=0.5\textwidth]{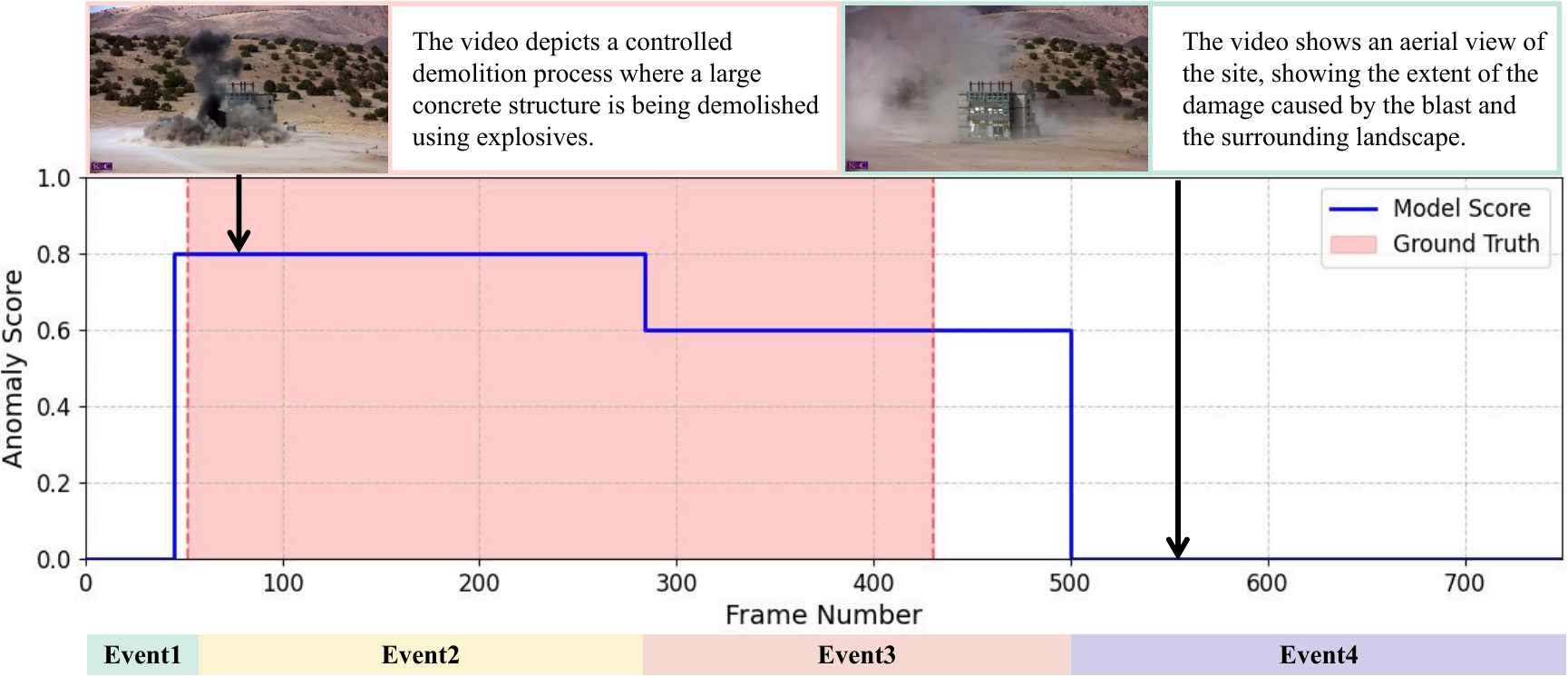}\label{fig:subfig9}}
  \subfloat[Normal Examples in XD-Violence Dataset]
  {\includegraphics[width=0.5\textwidth]{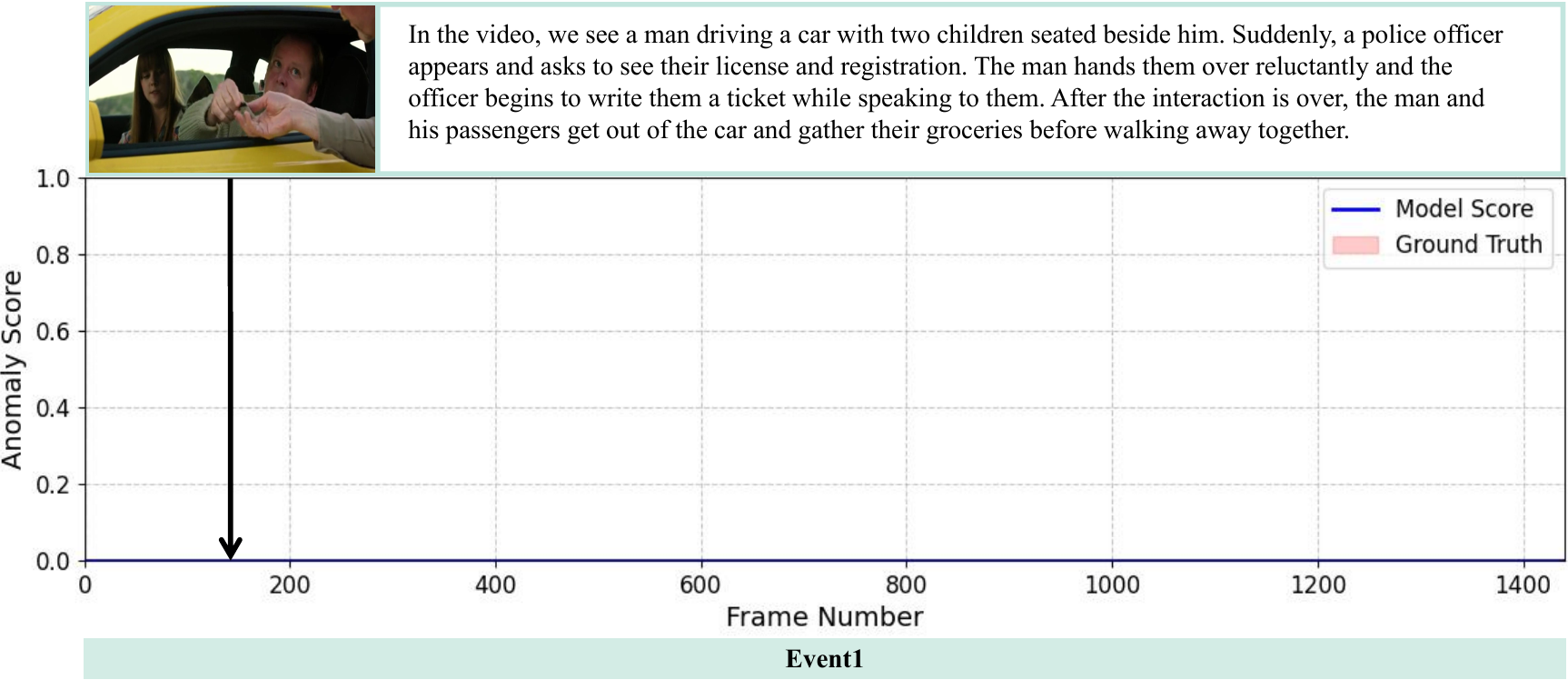}\label{fig:subfig10}}
 % \vspace{-0.3cm}
  \caption{
  % Visualization of detection results on UCF-Crime and XD-Violence datasets, showing cases where LAVAD failed to correctly identify anomalies. The display compares our event segmentation and MLLM anomaly scores against ground-truths.
  Visualization of normal and abnormal samples in the UCF-Crime and XD-Violence datasets. We visualized the samples where LAVAD failed to detect anomalies correctly. The figure presents event segmentation results along with the MLLM's anomaly frame scores, with comparisons to the ground-truth.
  }
   % \vspace{-0.2cm}
\end{figure*}

\noindent \textbf{Qualitative Results}
To intuitively demonstrate EventVAD's capabilities, we visually compare its performance against LAVAD by selecting failure cases from both the UCF-Crime and XD-Violence datasets for qualitative analysis. 

For anomalous samples, as shown in Fig.~\ref{fig:subfig7}, the video is a long sequence with over 2,500 frames, and the anomaly (robbery) is primarily concentrated in the lower-right corner of the footage. Therefore, most methods like LAVAD~\cite{zanella2024harnessing} struggle to detect such subtle anomalies. Although EventVAD's segmentation of event boundaries (start/end points) lacks precision, it effectively identifies the core segments of events. As shown in Fig.~\ref{fig:subfig9}, despite the visual prominence of the main subjects in the video, environmental interference can occasionally cause them to disappear from view. EventVAD segments videos into distinct events and leverages event consistency to enable precise anomaly localization within each event, effectively reducing false detections by the model.

For normal samples, as shown in Fig.~\ref{fig:subfig8} and Fig.~\ref{fig:subfig10}, Fig.~\ref{fig:subfig8} exhibits significant blurriness with poorly defined subjects, which poses substantial challenges for QA models like Blip~\cite{li2022blip,li2023blip} in performing accurate detection. Fig.~\ref{fig:subfig10} exceeds 1,400 frames in length, presenting significant detection challenges for LLMs~\cite{touvron2023llama,liu2023visual}. Using an event-aware approach, EventVAD optimizes video length for LLM input through event consistency. Even for lengthy videos, consistent events enable the LLM to achieve comprehensive anomaly detection by analyzing only the initial frames, thereby avoiding length-induced misjudgments.

\subsection{Ablation Study}
\label{sec:ablation-study}
% To validate the design of the EventVAD, we conduct ablation studies and evaluate each module with the UCF-Crime~\cite{sultani2018real} dataset. 
% We conduct a series of ablation studies on the UCF-Crime~\cite{sultani2018real} dataset.
% by gradually removing each module of EventVAD.
% \noindent \textbf{Analysis of proposed components}

\noindent \textbf{Efficiency-Accuracy Trade-off.}
Since the model parameters of the original LAVAD is 13 billion, we will explore the impact of model size on the framework's performance. As shown in Tab.~\ref{tab:parameter}, we first aligned the base model parameters of LAVAD and EventVAD. When both base models used Llama2~\cite{touvron2023llama}, we found that increasing LAVAD's model parameters could slightly enhance its anomaly detection capability. Although EventVAD also showed some improvement with larger model parameters, the gain was less pronounced compared to LAVAD. When using the 7B-parameter Video-Llama2~\cite{cheng2024videollama}, EventVAD achieved an AUC of 82.03\%, which is 3.69\% higher than the 13B-parameter LAVAD baseline.This validates the effectiveness of our event-centric architecture, which divides long videos into coherent segments to improve LLM comprehension, thus enabling reduced reliance on parametric scaling.
% The results on the XD-Violence dataset can be seen in Appendix~\ref{abl}.

\begin{table}[t]
    \centering
    \resizebox{0.95\linewidth}{!}{
\begin{tabular}{cccc}
\toprule
  \textbf{Method}       & \textbf{Foundational Model} & \textbf{Parameters} & \textbf{AUC (\%)}   \\ 
\midrule
LAVAD \cite{zanella2024harnessing}   & \multirow{2}{*}{Llama2-7b}& \multirow{2}{*}{7B} & 68.58 \\
\textbf{EventVAD (Ours)} & & & \textbf{78.39} \\
\midrule
LAVAD \cite{zanella2024harnessing} & \multirow{2}{*}{Llama2-13b}&\multirow{2}{*}{13B}& 78.33 \\ 
\textbf{EventVAD (Ours)} & & & \textbf{80.12} \\
\midrule
LAVAD \cite{zanella2024harnessing}    & \multirow{2}{*}{Video-Llama2} & \multirow{2}{*}{7B}& 73.28 \\
\textbf{EventVAD (Ours)} & & & \textbf{82.03} \\
\bottomrule
\end{tabular}}
    \caption{Results of scaling up the MLLMs
    }
    \label{tab:parameter}
     \vspace{-0.8cm}
\end{table}

\begin{table}[t]\small
\begin{tabular}{cccccc}
\toprule
   & $\alpha = 0$ & $\alpha = 0.25$ &$\alpha = 0.5$ &$\alpha = 0.75$ &$\alpha = 1$\\ \midrule
$\gamma =0.0$ &77.92    &77.13    &78.47    &78.91    &77.34    \\
$\gamma =0.2$  &78.22    &78.85    &79.48    &80.56    &77.93    \\
$\gamma =0.4$  &78.95    &79.12    &79.95    &81.78    &79.46    \\
$\gamma =0.6$ &79.89    &80.67    &81.57    &\textbf{\underline{82.03}}    &79.69    \\
$\gamma =0.8$  &78.92    &79.54    &80.38    &81.34    &79.62    \\
$\gamma =1.0$ &77.12    &78.81    &78.75    &80.97    &78.97    \\ \bottomrule
\end{tabular}
 \caption{Ablation Studies of Hyperparameters.
 % on EventVAD Performance. The experimental results demonstrate that our hyperparameter selection is reasonably well-justified.
 }
    \label{tab:hyparameter}
     \vspace{-0.8cm}
\end{table}
\noindent \textbf{Impact of Hyperparameters on Model Performance.}
 EventVAD's core hyperparameters - the temporal decay coefficient $\gamma$ and semantic-motion fusion coefficient $\alpha$ - critically govern model behavior. We present a systematic analysis of their impacts.
 
As shown in Table~\ref{tab:hyparameter}, the temporal decay factor $γ$ and the semantic-motion fusion coefficient $α$ have a serious impact on model behavior through their coordinated control of spatiotemporal feature integration. With the increase of $α$ and $γ$, it is initially possible for the framework to improve the quality of segmentation by enhancing the semantic correlation between frames while maintaining appropriate temporal continuity constraints.
However, too high parameter values will lead to performance degradation, which will make the model overly sensitive to semantic changes, generate fragmented fragments, and thus disrupt event continuity. These ultra-short fragments prevent MLLMs from establishing a full semantic context, resulting in hallucinogenic anomalous scoring.

To maintain a good semantic correlation between frames while having a certain time continuity constraint, we choose $\alpha=0.75$ and $\gamma=0.6$ for segmentation.

\begin{table}[t]
\small
\begin{tabular}{ccc|cc}
\toprule
\textbf{RAFT} & \textbf{GANP} & \textbf{Thinking} & \textbf{AUC (\%)} & $\Delta$ \textbf{(\%)} \\ \midrule
\faTimes    & \faTimes         & \faTimes          & 73.93  & -  \\
\faTimes      & \Checkmark         & \faTimes        & 77.81  & +3.88   \\
\Checkmark     & \faTimes         & \faTimes         & 75.35  & +1.42  \\
\Checkmark   & \Checkmark         & \faTimes   & 80.14  & +6.21  \\ 
\faTimes     & \faTimes         & \Checkmark          & 75.64  & +1.71   \\
\faTimes       & \Checkmark         & \Checkmark        & 79.97  & +6.04  \\
\Checkmark     & \faTimes         & \Checkmark         & 76.86  & +2.93  \\
\Checkmark   & \Checkmark         & \Checkmark   & 82.03  & +8.10  \\ 
\bottomrule
\end{tabular}
\caption{Ablation Studies of Key Modules on EventVAD.
Through ablation studies, we validate their effectiveness and identify the most critical ones.
}
\label{tab:ab3}
 \vspace{-2em}
\end{table}

\noindent \textbf{Key Module Ablation.}
For EventVAD, the Optical Flow Module, the Graph Attention propagation module, and the video content description before MLLMs scoring provide the main contribution to the model's performance. To systematically evaluate each component's contribution, we conducted modular ablation studies focusing on three key aspects: (1) the inclusion of RAFT-based optical flow features during encoding, (2) the use of graph attention network propagation (GANP) following dynamic graph construction, and (3) the implementation of deliberative reasoning in MLLM outputs before anomaly scoring. We further analyzed cross-component interactions to explore potential synergies.

As shown in Table~\ref{tab:ab3}, when the original model is combined with the RAFT optical flow, the detection AUC increases by 2\%, while with the help of graph attention propagation, the detection AUC of the model is further increased by about 4\%. This is a significant improvement over static graphs and solves the problem of long graph neural network training times. Although RAFT optical flow cannot directly enhance the capabilities of the Dynamic Graph or Graph Attention Propagation modules, it can enhance the modeling capabilities of the Dynamic Graph through temporal feature matching, thereby improving its boundary detection performance. In addition, the quantitative experiments show the structured MLLM output can be specified, so that it can summarize the video content before outputting the abnormal score, and this process of letting the model think can better help him understand these small fragments and give a more reasonable score.
% \begin{table}[]
%     \centering
% \begin{tabular}{cccc}
% \toprule
%   \textbf{Method}       & \textbf{Foundational Model} & \textbf{Parameters} & \textbf{AUC (\%)}   \\ \midrule
% EventVAD & Video-Llama2~\cite{cheng2024videollama}       & 7B         & 82.03 \\
% LAVAD    & Video-Llama2~\cite{cheng2024videollama}       & 7B         & 73.28 \\
% EventVAD & Llama2-7b~\cite{touvron2023llama}             & 7B         & 70.39 \\
% LAVAD    & Llama2-7b~\cite{touvron2023llama}             & 7B         & 68.58 \\
% EventVAD & Llama2-13b~\cite{touvron2023llama}             & 13B        & 81.92 \\
% LAVAD    & Llama2-13b~\cite{touvron2023llama}             & 13B        & 80.28 \\ \bottomrule
% \end{tabular}
%     \caption{Caption}
%     \label{tab:my_label}
% \end{table}

\section{Conclusion}
In this paper, we propose a novel training-free event-aware video anomaly detection framework, EventVAD. EventVAD addresses the challenge of localizing fine-grained visual transitions and diverse events in long videos by segmenting events in videos. 
EventVAD achieves more accurate event boundary detection and events division by introducing optical flow matching and dynamic graph architectures so that MLLMs can score anomalies accurately within events. 
We evaluated EventVAD on both UCF-Crime and XD-Violence datasets. 
Experimental results demonstrate that EventVAD can segment events accurately in videos and achieves SOTA in video anomaly detection on both datasets in the training-free setting with 7B MLLMs compared to baseline 13B or larger MLLMs. 
In the future, we will explore more efficient techniques, such as pruning and quantization, to facilitate the deployment of our method in resource-constrained scenarios.
% The results also demonstrate that EventVAD outperforms unsupervised and one-class settings, which demonstrates that EventVAD exhibits strong generalization capabilities.
% So, this work makes a substantial contribution to the field by introducing a flexible and robust framework for video anomaly detection.
\section*{Acknowledgements}
This research was supported by funding from the Start-up Package, School of EEECS, Queen's University Belfast (D8203EEC)

% \clearpage
\bibliographystyle{ACM-Reference-Format}
\bibliography{sample-base}


\begin{thebibliography}{75}

%%% the Web address.

\ifx \showCODEN    \undefined \def \showCODEN     #1{\unskip}     \fi
\ifx \showDOI      \undefined \def \showDOI       #1{#1}\fi
\ifx \showISBNx    \undefined \def \showISBNx     #1{\unskip}     \fi
\ifx \showISBNxiii \undefined \def \showISBNxiii  #1{\unskip}     \fi
\ifx \showISSN     \undefined \def \showISSN      #1{\unskip}     \fi
\ifx \showLCCN     \undefined \def \showLCCN      #1{\unskip}     \fi
\ifx \shownote     \undefined \def \shownote      #1{#1}          \fi
\ifx \showarticletitle \undefined \def \showarticletitle #1{#1}   \fi
\ifx \showURL      \undefined \def \showURL       {\relax}        \fi
% invisible to TeX
\providecommand\bibfield[2]{#2}
\providecommand\bibinfo[2]{#2}
\providecommand\natexlab[1]{#1}
\providecommand\showeprint[2][]{arXiv:#2}

\bibitem[Alayrac et~al\mbox{.}(2022)]%
        {alayrac2022flamingo}
\bibfield{author}{\bibinfo{person}{Jean-Baptiste Alayrac}, \bibinfo{person}{Jeff Donahue}, \bibinfo{person}{Pauline Luc}, \bibinfo{person}{Antoine Miech}, \bibinfo{person}{Iain Barr}, \bibinfo{person}{Yana Hasson}, \bibinfo{person}{Karel Lenc}, \bibinfo{person}{Arthur Mensch}, \bibinfo{person}{Katherine Millican}, \bibinfo{person}{Malcolm Reynolds}, {et~al\mbox{.}}} \bibinfo{year}{2022}\natexlab{}.
\newblock \showarticletitle{Flamingo: a visual language model for few-shot learning}.
\newblock \bibinfo{journal}{\emph{Advances in neural information processing systems}}  \bibinfo{volume}{35} (\bibinfo{year}{2022}), \bibinfo{pages}{23716--23736}.
\newblock


\bibitem[Ataallah et~al\mbox{.}(2024)]%
        {ataallah2024minigpt4}
\bibfield{author}{\bibinfo{person}{Kirolos Ataallah}, \bibinfo{person}{Xiaoqian Shen}, \bibinfo{person}{Eslam Abdelrahman}, \bibinfo{person}{Essam Sleiman}, \bibinfo{person}{Deyao Zhu}, \bibinfo{person}{Jian Ding}, {and} \bibinfo{person}{Mohamed Elhoseiny}.} \bibinfo{year}{2024}\natexlab{}.
\newblock \showarticletitle{Minigpt4-video: Advancing multimodal llms for video understanding with interleaved visual-textual tokens}.
\newblock \bibinfo{journal}{\emph{arXiv preprint arXiv:2404.03413}} (\bibinfo{year}{2024}).
\newblock


\bibitem[Chen et~al\mbox{.}(2023b)]%
        {chen2023shikra}
\bibfield{author}{\bibinfo{person}{Keqin Chen}, \bibinfo{person}{Zhao Zhang}, \bibinfo{person}{Weili Zeng}, \bibinfo{person}{Richong Zhang}, \bibinfo{person}{Feng Zhu}, {and} \bibinfo{person}{Rui Zhao}.} \bibinfo{year}{2023}\natexlab{b}.
\newblock \showarticletitle{Shikra: Unleashing multimodal llm's referential dialogue magic}.
\newblock \bibinfo{journal}{\emph{arXiv preprint arXiv:2306.15195}} (\bibinfo{year}{2023}).
\newblock


\bibitem[Chen et~al\mbox{.}(2023a)]%
        {chen2023mgfn}
\bibfield{author}{\bibinfo{person}{Yingxian Chen}, \bibinfo{person}{Zhengzhe Liu}, \bibinfo{person}{Baoheng Zhang}, \bibinfo{person}{Wilton Fok}, \bibinfo{person}{Xiaojuan Qi}, {and} \bibinfo{person}{Yik-Chung Wu}.} \bibinfo{year}{2023}\natexlab{a}.
\newblock \showarticletitle{Mgfn: Magnitude-contrastive glance-and-focus network for weakly-supervised video anomaly detection}. In \bibinfo{booktitle}{\emph{AAAI}}.
\newblock


\bibitem[Cheng et~al\mbox{.}(2024)]%
        {cheng2024videollama}
\bibfield{author}{\bibinfo{person}{Zesen Cheng}, \bibinfo{person}{Sicong Leng}, \bibinfo{person}{Hang Zhang}, \bibinfo{person}{Yifei Xin}, \bibinfo{person}{Xin Li}, \bibinfo{person}{Guanzheng Chen}, \bibinfo{person}{Yongxin Zhu}, \bibinfo{person}{Wenqi Zhang}, \bibinfo{person}{Ziyang Luo}, \bibinfo{person}{Deli Zhao}, {et~al\mbox{.}}} \bibinfo{year}{2024}\natexlab{}.
\newblock \showarticletitle{Videollama 2: Advancing spatial-temporal modeling and audio understanding in video-llms}.
\newblock \bibinfo{journal}{\emph{arXiv preprint arXiv:2406.07476}} (\bibinfo{year}{2024}).
\newblock


\bibitem[Dev et~al\mbox{.}(2025)]%
        {dev2025mcanet}
\bibfield{author}{\bibinfo{person}{Prabhu~Prasad Dev}, \bibinfo{person}{Raju Hazari}, {and} \bibinfo{person}{Pranesh Das}.} \bibinfo{year}{2025}\natexlab{}.
\newblock \showarticletitle{MCANet: Multimodal Caption Aware Training-Free Video Anomaly Detection via Large Language Model}. In \bibinfo{booktitle}{\emph{International Conference on Pattern Recognition}}. Springer, \bibinfo{pages}{362--379}.
\newblock


\bibitem[Girdhar et~al\mbox{.}(2023)]%
        {girdhar2023imagebind}
\bibfield{author}{\bibinfo{person}{Rohit Girdhar}, \bibinfo{person}{Alaaeldin El-Nouby}, \bibinfo{person}{Zhuang Liu}, \bibinfo{person}{Mannat Singh}, \bibinfo{person}{Kalyan~Vasudev Alwala}, \bibinfo{person}{Armand Joulin}, {and} \bibinfo{person}{Ishan Misra}.} \bibinfo{year}{2023}\natexlab{}.
\newblock \showarticletitle{Imagebind: One embedding space to bind them all}. In \bibinfo{booktitle}{\emph{CVPR}}.
\newblock


\bibitem[Grattafiori et~al\mbox{.}(2024)]%
        {grattafiori2024llama}
\bibfield{author}{\bibinfo{person}{Aaron Grattafiori}, \bibinfo{person}{Abhimanyu Dubey}, \bibinfo{person}{Abhinav Jauhri}, \bibinfo{person}{Abhinav Pandey}, \bibinfo{person}{Abhishek Kadian}, \bibinfo{person}{Ahmad Al-Dahle}, \bibinfo{person}{Aiesha Letman}, \bibinfo{person}{Akhil Mathur}, \bibinfo{person}{Alan Schelten}, \bibinfo{person}{Alex Vaughan}, {et~al\mbox{.}}} \bibinfo{year}{2024}\natexlab{}.
\newblock \showarticletitle{The llama 3 herd of models}.
\newblock \bibinfo{journal}{\emph{arXiv preprint arXiv:2407.21783}} (\bibinfo{year}{2024}).
\newblock


\bibitem[Hasan et~al\mbox{.}(2016)]%
        {hasan2016learning}
\bibfield{author}{\bibinfo{person}{Mahmudul Hasan}, \bibinfo{person}{Jonghyun Choi}, \bibinfo{person}{Jan Neumann}, \bibinfo{person}{Amit~K Roy-Chowdhury}, {and} \bibinfo{person}{Larry~S Davis}.} \bibinfo{year}{2016}\natexlab{}.
\newblock \showarticletitle{Learning temporal regularity in video sequences}. In \bibinfo{booktitle}{\emph{CVPR}}.
\newblock


\bibitem[Hong et~al\mbox{.}(2024)]%
        {hong2024cogvlm2}
\bibfield{author}{\bibinfo{person}{Wenyi Hong}, \bibinfo{person}{Weihan Wang}, \bibinfo{person}{Ming Ding}, \bibinfo{person}{Wenmeng Yu}, \bibinfo{person}{Qingsong Lv}, \bibinfo{person}{Yan Wang}, \bibinfo{person}{Yean Cheng}, \bibinfo{person}{Shiyu Huang}, \bibinfo{person}{Junhui Ji}, \bibinfo{person}{Zhao Xue}, {et~al\mbox{.}}} \bibinfo{year}{2024}\natexlab{}.
\newblock \showarticletitle{Cogvlm2: Visual language models for image and video understanding}.
\newblock \bibinfo{journal}{\emph{arXiv preprint arXiv:2408.16500}} (\bibinfo{year}{2024}).
\newblock


\bibitem[Huang et~al\mbox{.}(2024)]%
        {huang2024vtimellm}
\bibfield{author}{\bibinfo{person}{Bin Huang}, \bibinfo{person}{Xin Wang}, \bibinfo{person}{Hong Chen}, \bibinfo{person}{Zihan Song}, {and} \bibinfo{person}{Wenwu Zhu}.} \bibinfo{year}{2024}\natexlab{}.
\newblock \showarticletitle{Vtimellm: Empower llm to grasp video moments}. In \bibinfo{booktitle}{\emph{Proceedings of the IEEE/CVF Conference on Computer Vision and Pattern Recognition}}. \bibinfo{pages}{14271--14280}.
\newblock


\bibitem[Jin et~al\mbox{.}(2024)]%
        {jin2024video}
\bibfield{author}{\bibinfo{person}{Yang Jin}, \bibinfo{person}{Zhicheng Sun}, \bibinfo{person}{Kun Xu}, \bibinfo{person}{Liwei Chen}, \bibinfo{person}{Hao Jiang}, \bibinfo{person}{Quzhe Huang}, \bibinfo{person}{Chengru Song}, \bibinfo{person}{Yuliang Liu}, \bibinfo{person}{Di Zhang}, \bibinfo{person}{Yang Song}, {et~al\mbox{.}}} \bibinfo{year}{2024}\natexlab{}.
\newblock \showarticletitle{Video-lavit: Unified video-language pre-training with decoupled visual-motional tokenization}.
\newblock \bibinfo{journal}{\emph{arXiv preprint arXiv:2402.03161}} (\bibinfo{year}{2024}).
\newblock


\bibitem[Jin et~al\mbox{.}(2023)]%
        {jin2023unified}
\bibfield{author}{\bibinfo{person}{Yang Jin}, \bibinfo{person}{Kun Xu}, \bibinfo{person}{Liwei Chen}, \bibinfo{person}{Chao Liao}, \bibinfo{person}{Jianchao Tan}, \bibinfo{person}{Quzhe Huang}, \bibinfo{person}{Bin Chen}, \bibinfo{person}{Chenyi Lei}, \bibinfo{person}{An Liu}, \bibinfo{person}{Chengru Song}, {et~al\mbox{.}}} \bibinfo{year}{2023}\natexlab{}.
\newblock \showarticletitle{Unified language-vision pretraining in llm with dynamic discrete visual tokenization}.
\newblock \bibinfo{journal}{\emph{arXiv preprint arXiv:2309.04669}} (\bibinfo{year}{2023}).
\newblock


\bibitem[Joo et~al\mbox{.}(2023)]%
        {joo2023clip}
\bibfield{author}{\bibinfo{person}{Hyekang~Kevin Joo}, \bibinfo{person}{Khoa Vo}, \bibinfo{person}{Kashu Yamazaki}, {and} \bibinfo{person}{Ngan Le}.} \bibinfo{year}{2023}\natexlab{}.
\newblock \showarticletitle{CLIP-TSA: CLIP-Assisted Temporal Self-Attention for Weakly-Supervised Video Anomaly Detection}. In \bibinfo{booktitle}{\emph{ICIP}}.
\newblock


\bibitem[Leys et~al\mbox{.}(2013)]%
        {leys2013detecting}
\bibfield{author}{\bibinfo{person}{Christophe Leys}, \bibinfo{person}{Christophe Ley}, \bibinfo{person}{Olivier Klein}, \bibinfo{person}{Philippe Bernard}, {and} \bibinfo{person}{Laurent Licata}.} \bibinfo{year}{2013}\natexlab{}.
\newblock \showarticletitle{Detecting outliers: Do not use standard deviation around the mean, use absolute deviation around the median}.
\newblock \bibinfo{journal}{\emph{Journal of experimental social psychology}} \bibinfo{volume}{49}, \bibinfo{number}{4} (\bibinfo{year}{2013}), \bibinfo{pages}{764--766}.
\newblock


\bibitem[Li et~al\mbox{.}(2022a)]%
        {li2022scale}
\bibfield{author}{\bibinfo{person}{Guoqiu Li}, \bibinfo{person}{Guanxiong Cai}, \bibinfo{person}{Xingyu Zeng}, {and} \bibinfo{person}{Rui Zhao}.} \bibinfo{year}{2022}\natexlab{a}.
\newblock \showarticletitle{Scale-Aware Spatio-Temporal Relation Learning for Video Anomaly Detection}. In \bibinfo{booktitle}{\emph{ECCV}}.
\newblock


\bibitem[Li et~al\mbox{.}(2023)]%
        {li2023blip}
\bibfield{author}{\bibinfo{person}{Junnan Li}, \bibinfo{person}{Dongxu Li}, \bibinfo{person}{Silvio Savarese}, {and} \bibinfo{person}{Steven Hoi}.} \bibinfo{year}{2023}\natexlab{}.
\newblock \showarticletitle{Blip-2: Bootstrapping language-image pre-training with frozen image encoders and large language models}. In \bibinfo{booktitle}{\emph{ICML}}.
\newblock


\bibitem[Li et~al\mbox{.}(2022b)]%
        {li2022blip}
\bibfield{author}{\bibinfo{person}{Junnan Li}, \bibinfo{person}{Dongxu Li}, \bibinfo{person}{Caiming Xiong}, {and} \bibinfo{person}{Steven Hoi}.} \bibinfo{year}{2022}\natexlab{b}.
\newblock \showarticletitle{Blip: Bootstrapping language-image pre-training for unified vision-language understanding and generation}. In \bibinfo{booktitle}{\emph{International conference on machine learning}}. PMLR, \bibinfo{pages}{12888--12900}.
\newblock


\bibitem[Li et~al\mbox{.}(2024)]%
        {li2024mvbench}
\bibfield{author}{\bibinfo{person}{Kunchang Li}, \bibinfo{person}{Yali Wang}, \bibinfo{person}{Yinan He}, \bibinfo{person}{Yizhuo Li}, \bibinfo{person}{Yi Wang}, \bibinfo{person}{Yi Liu}, \bibinfo{person}{Zun Wang}, \bibinfo{person}{Jilan Xu}, \bibinfo{person}{Guo Chen}, \bibinfo{person}{Ping Luo}, {et~al\mbox{.}}} \bibinfo{year}{2024}\natexlab{}.
\newblock \showarticletitle{Mvbench: A comprehensive multi-modal video understanding benchmark}. In \bibinfo{booktitle}{\emph{Proceedings of the IEEE/CVF Conference on Computer Vision and Pattern Recognition}}. \bibinfo{pages}{22195--22206}.
\newblock


\bibitem[Li et~al\mbox{.}(2022c)]%
        {li2022self}
\bibfield{author}{\bibinfo{person}{Shuo Li}, \bibinfo{person}{Fang Liu}, {and} \bibinfo{person}{Licheng Jiao}.} \bibinfo{year}{2022}\natexlab{c}.
\newblock \showarticletitle{Self-training multi-sequence learning with transformer for weakly supervised video anomaly detection}. In \bibinfo{booktitle}{\emph{AAAI}}.
\newblock


\bibitem[Liao et~al\mbox{.}(2025)]%
        {liao2025gm}
\bibfield{author}{\bibinfo{person}{Minwen Liao}, \bibinfo{person}{Hao~Bo Dong}, \bibinfo{person}{Xinyi Wang}, \bibinfo{person}{Ziyang Yan}, {and} \bibinfo{person}{Yihua Shao}.} \bibinfo{year}{2025}\natexlab{}.
\newblock \showarticletitle{GM-MoE: Low-Light Enhancement with Gated-Mechanism Mixture-of-Experts}.
\newblock \bibinfo{journal}{\emph{arXiv preprint arXiv:2503.07417}} (\bibinfo{year}{2025}).
\newblock


\bibitem[Lin et~al\mbox{.}(2023)]%
        {lin2023video}
\bibfield{author}{\bibinfo{person}{Bin Lin}, \bibinfo{person}{Yang Ye}, \bibinfo{person}{Bin Zhu}, \bibinfo{person}{Jiaxi Cui}, \bibinfo{person}{Munan Ning}, \bibinfo{person}{Peng Jin}, {and} \bibinfo{person}{Li Yuan}.} \bibinfo{year}{2023}\natexlab{}.
\newblock \showarticletitle{Video-llava: Learning united visual representation by alignment before projection}.
\newblock \bibinfo{journal}{\emph{arXiv preprint arXiv:2311.10122}} (\bibinfo{year}{2023}).
\newblock


\bibitem[Liu et~al\mbox{.}(2023a)]%
        {liu2023improved}
\bibfield{author}{\bibinfo{person}{Haotian Liu}, \bibinfo{person}{Chunyuan Li}, \bibinfo{person}{Yuheng Li}, {and} \bibinfo{person}{Yong~Jae Lee}.} \bibinfo{year}{2023}\natexlab{a}.
\newblock \showarticletitle{Improved baselines with visual instruction tuning}.
\newblock \bibinfo{journal}{\emph{arXiv}} (\bibinfo{year}{2023}).
\newblock


\bibitem[Liu et~al\mbox{.}(2023b)]%
        {liu2023visual}
\bibfield{author}{\bibinfo{person}{Haotian Liu}, \bibinfo{person}{Chunyuan Li}, \bibinfo{person}{Qingyang Wu}, {and} \bibinfo{person}{Yong~Jae Lee}.} \bibinfo{year}{2023}\natexlab{b}.
\newblock \showarticletitle{Visual instruction tuning}.
\newblock \bibinfo{journal}{\emph{Advances in neural information processing systems}}  \bibinfo{volume}{36} (\bibinfo{year}{2023}), \bibinfo{pages}{34892--34916}.
\newblock


\bibitem[Liu et~al\mbox{.}(2024)]%
        {liu2024st}
\bibfield{author}{\bibinfo{person}{Ruyang Liu}, \bibinfo{person}{Chen Li}, \bibinfo{person}{Haoran Tang}, \bibinfo{person}{Yixiao Ge}, \bibinfo{person}{Ying Shan}, {and} \bibinfo{person}{Ge Li}.} \bibinfo{year}{2024}\natexlab{}.
\newblock \showarticletitle{St-llm: Large language models are effective temporal learners}. In \bibinfo{booktitle}{\emph{European Conference on Computer Vision}}. Springer, \bibinfo{pages}{1--18}.
\newblock


\bibitem[Long et~al\mbox{.}(2025)]%
        {long2025retrieval}
\bibfield{author}{\bibinfo{person}{Xinwei Long}, \bibinfo{person}{Zhiyuan Ma}, \bibinfo{person}{Ermo Hua}, \bibinfo{person}{Kaiyan Zhang}, \bibinfo{person}{Biqing Qi}, {and} \bibinfo{person}{Bowen Zhou}.} \bibinfo{year}{2025}\natexlab{}.
\newblock \showarticletitle{Retrieval-Augmented Visual Question Answering via Built-in Autoregressive Search Engines}. In \bibinfo{booktitle}{\emph{Proceedings of the AAAI Conference on Artificial Intelligence}}, Vol.~\bibinfo{volume}{39}. \bibinfo{pages}{24723--24731}.
\newblock


\bibitem[Long et~al\mbox{.}(2021)]%
        {long2021position}
\bibfield{author}{\bibinfo{person}{Xinwei Long}, \bibinfo{person}{Shuzi Niu}, {and} \bibinfo{person}{Yucheng Li}.} \bibinfo{year}{2021}\natexlab{}.
\newblock \showarticletitle{Position enhanced mention graph attention network for dialogue relation extraction}. In \bibinfo{booktitle}{\emph{Proceedings of the 44th International ACM SIGIR Conference on Research and Development in Information Retrieval}}. \bibinfo{pages}{1985--1989}.
\newblock


\bibitem[Long et~al\mbox{.}(2024a)]%
        {long2024generative}
\bibfield{author}{\bibinfo{person}{Xinwei Long}, \bibinfo{person}{Jiali Zeng}, \bibinfo{person}{Fandong Meng}, \bibinfo{person}{Zhiyuan Ma}, \bibinfo{person}{Kaiyan Zhang}, \bibinfo{person}{Bowen Zhou}, {and} \bibinfo{person}{Jie Zhou}.} \bibinfo{year}{2024}\natexlab{a}.
\newblock \showarticletitle{Generative multi-modal knowledge retrieval with large language models}. In \bibinfo{booktitle}{\emph{Proceedings of the AAAI Conference on Artificial Intelligence}}, Vol.~\bibinfo{volume}{38}. \bibinfo{pages}{18733--18741}.
\newblock


\bibitem[Long et~al\mbox{.}(2024b)]%
        {long2024trust}
\bibfield{author}{\bibinfo{person}{Xinwei Long}, \bibinfo{person}{Jiali Zeng}, \bibinfo{person}{Fandong Meng}, \bibinfo{person}{Jie Zhou}, {and} \bibinfo{person}{Bowen Zhou}.} \bibinfo{year}{2024}\natexlab{b}.
\newblock \showarticletitle{Trust in internal or external knowledge? generative multi-modal entity linking with knowledge retriever}. In \bibinfo{booktitle}{\emph{Findings of the Association for Computational Linguistics ACL 2024}}. \bibinfo{pages}{7559--7569}.
\newblock


\bibitem[Lu et~al\mbox{.}(2013)]%
        {lu2013abnormal}
\bibfield{author}{\bibinfo{person}{Cewu Lu}, \bibinfo{person}{Jianping Shi}, {and} \bibinfo{person}{Jiaya Jia}.} \bibinfo{year}{2013}\natexlab{}.
\newblock \showarticletitle{Abnormal event detection at 150 fps in matlab}. In \bibinfo{booktitle}{\emph{ICCV}}.
\newblock


\bibitem[Maaz et~al\mbox{.}(2023)]%
        {maaz2023video}
\bibfield{author}{\bibinfo{person}{Muhammad Maaz}, \bibinfo{person}{Hanoona Rasheed}, \bibinfo{person}{Salman Khan}, {and} \bibinfo{person}{Fahad~Shahbaz Khan}.} \bibinfo{year}{2023}\natexlab{}.
\newblock \showarticletitle{Video-chatgpt: Towards detailed video understanding via large vision and language models}.
\newblock \bibinfo{journal}{\emph{arXiv preprint arXiv:2306.05424}} (\bibinfo{year}{2023}).
\newblock


\bibitem[Nayak et~al\mbox{.}(2021)]%
        {nayak2021comprehensive}
\bibfield{author}{\bibinfo{person}{Rashmiranjan Nayak}, \bibinfo{person}{Umesh~Chandra Pati}, {and} \bibinfo{person}{Santos~Kumar Das}.} \bibinfo{year}{2021}\natexlab{}.
\newblock \showarticletitle{A comprehensive review on deep learning-based methods for video anomaly detection}.
\newblock \bibinfo{journal}{\emph{Image and Vision Computing}} (\bibinfo{year}{2021}).
\newblock


\bibitem[Radford et~al\mbox{.}(2021)]%
        {radford2021learning}
\bibfield{author}{\bibinfo{person}{Alec Radford}, \bibinfo{person}{Jong~Wook Kim}, \bibinfo{person}{Chris Hallacy}, \bibinfo{person}{Aditya Ramesh}, \bibinfo{person}{Gabriel Goh}, \bibinfo{person}{Sandhini Agarwal}, \bibinfo{person}{Girish Sastry}, \bibinfo{person}{Amanda Askell}, \bibinfo{person}{Pamela Mishkin}, \bibinfo{person}{Jack Clark}, {et~al\mbox{.}}} \bibinfo{year}{2021}\natexlab{}.
\newblock \showarticletitle{Learning transferable visual models from natural language supervision}. In \bibinfo{booktitle}{\emph{International conference on machine learning}}. PmLR, \bibinfo{pages}{8748--8763}.
\newblock


\bibitem[Ramachandra et~al\mbox{.}(2020)]%
        {ramachandra2020survey}
\bibfield{author}{\bibinfo{person}{Bharathkumar Ramachandra}, \bibinfo{person}{Michael~J Jones}, {and} \bibinfo{person}{Ranga~Raju Vatsavai}.} \bibinfo{year}{2020}\natexlab{}.
\newblock \showarticletitle{A survey of single-scene video anomaly detection}.
\newblock \bibinfo{journal}{\emph{IEEE transactions on pattern analysis and machine intelligence}} \bibinfo{volume}{44}, \bibinfo{number}{5} (\bibinfo{year}{2020}), \bibinfo{pages}{2293--2312}.
\newblock


\bibitem[Remondino et~al\mbox{.}(2023)]%
        {remondino2023critical}
\bibfield{author}{\bibinfo{person}{Fabio Remondino}, \bibinfo{person}{Ali Karami}, \bibinfo{person}{Ziyang Yan}, \bibinfo{person}{Gabriele Mazzacca}, \bibinfo{person}{Simone Rigon}, {and} \bibinfo{person}{Rongjun Qin}.} \bibinfo{year}{2023}\natexlab{}.
\newblock \showarticletitle{A critical analysis of NeRF-based 3D reconstruction}.
\newblock \bibinfo{journal}{\emph{Remote Sensing}} \bibinfo{volume}{15}, \bibinfo{number}{14} (\bibinfo{year}{2023}), \bibinfo{pages}{3585}.
\newblock


\bibitem[Scarselli et~al\mbox{.}(2008)]%
        {scarselli2008graph}
\bibfield{author}{\bibinfo{person}{Franco Scarselli}, \bibinfo{person}{Marco Gori}, \bibinfo{person}{Ah~Chung Tsoi}, \bibinfo{person}{Markus Hagenbuchner}, {and} \bibinfo{person}{Gabriele Monfardini}.} \bibinfo{year}{2008}\natexlab{}.
\newblock \showarticletitle{The graph neural network model}.
\newblock \bibinfo{journal}{\emph{IEEE transactions on neural networks}} \bibinfo{volume}{20}, \bibinfo{number}{1} (\bibinfo{year}{2008}), \bibinfo{pages}{61--80}.
\newblock


\bibitem[Shao et~al\mbox{.}(2024a)]%
        {shao2024gwq}
\bibfield{author}{\bibinfo{person}{Yihua Shao}, \bibinfo{person}{Siyu Liang}, \bibinfo{person}{Zijian Ling}, \bibinfo{person}{Minxi Yan}, \bibinfo{person}{Haiyang Liu}, \bibinfo{person}{Siyu Chen}, \bibinfo{person}{Ziyang Yan}, \bibinfo{person}{Chenyu Zhang}, \bibinfo{person}{Haotong Qin}, \bibinfo{person}{Michele Magno}, {et~al\mbox{.}}} \bibinfo{year}{2024}\natexlab{a}.
\newblock \showarticletitle{GWQ: Gradient-Aware Weight Quantization for Large Language Models}.
\newblock \bibinfo{journal}{\emph{arXiv preprint arXiv:2411.00850}} (\bibinfo{year}{2024}).
\newblock


\bibitem[Shao et~al\mbox{.}(2025a)]%
        {shao2025tr}
\bibfield{author}{\bibinfo{person}{Yihua Shao}, \bibinfo{person}{Deyang Lin}, \bibinfo{person}{Fanhu Zeng}, \bibinfo{person}{Minxi Yan}, \bibinfo{person}{Muyang Zhang}, \bibinfo{person}{Siyu Chen}, \bibinfo{person}{Yuxuan Fan}, \bibinfo{person}{Ziyang Yan}, \bibinfo{person}{Haozhe Wang}, \bibinfo{person}{Jingcai Guo}, {et~al\mbox{.}}} \bibinfo{year}{2025}\natexlab{a}.
\newblock \showarticletitle{TR-DQ: Time-Rotation Diffusion Quantization}.
\newblock \bibinfo{journal}{\emph{arXiv preprint arXiv:2503.06564}} (\bibinfo{year}{2025}).
\newblock


\bibitem[Shao et~al\mbox{.}(2024b)]%
        {shao2024accidentblip}
\bibfield{author}{\bibinfo{person}{Yihua Shao}, \bibinfo{person}{Yeling Xu}, \bibinfo{person}{Xinwei Long}, \bibinfo{person}{Siyu Chen}, \bibinfo{person}{Ziyang Yan}, \bibinfo{person}{Yang Yang}, \bibinfo{person}{Haoting Liu}, \bibinfo{person}{Yan Wang}, \bibinfo{person}{Hao Tang}, {and} \bibinfo{person}{Zhen Lei}.} \bibinfo{year}{2024}\natexlab{b}.
\newblock \showarticletitle{AccidentBlip: Agent of Accident Warning based on MA-former}.
\newblock \bibinfo{journal}{\emph{arXiv preprint arXiv:2404.12149}} (\bibinfo{year}{2024}).
\newblock


\bibitem[Shao et~al\mbox{.}(2025b)]%
        {shao2025context}
\bibfield{author}{\bibinfo{person}{Yihua Shao}, \bibinfo{person}{Minxi Yan}, \bibinfo{person}{Yang Liu}, \bibinfo{person}{Siyu Chen}, \bibinfo{person}{Wenjie Chen}, \bibinfo{person}{Xinwei Long}, \bibinfo{person}{Ziyang Yan}, \bibinfo{person}{Lei Li}, \bibinfo{person}{Chenyu Zhang}, \bibinfo{person}{Nicu Sebe}, {et~al\mbox{.}}} \bibinfo{year}{2025}\natexlab{b}.
\newblock \showarticletitle{In-Context Meta LoRA Generation}.
\newblock \bibinfo{journal}{\emph{arXiv preprint arXiv:2501.17635}} (\bibinfo{year}{2025}).
\newblock


\bibitem[Sohrab et~al\mbox{.}(2018)]%
        {sohrab2018subspace}
\bibfield{author}{\bibinfo{person}{Fahad Sohrab}, \bibinfo{person}{Jenni Raitoharju}, \bibinfo{person}{Moncef Gabbouj}, {and} \bibinfo{person}{Alexandros Iosifidis}.} \bibinfo{year}{2018}\natexlab{}.
\newblock \showarticletitle{Subspace support vector data description}. In \bibinfo{booktitle}{\emph{ICPR}}.
\newblock


\bibitem[Su et~al\mbox{.}(2023)]%
        {su2023pandagpt}
\bibfield{author}{\bibinfo{person}{Yixuan Su}, \bibinfo{person}{Tian Lan}, \bibinfo{person}{Huayang Li}, \bibinfo{person}{Jialu Xu}, \bibinfo{person}{Yan Wang}, {and} \bibinfo{person}{Deng Cai}.} \bibinfo{year}{2023}\natexlab{}.
\newblock \showarticletitle{Pandagpt: One model to instruction-follow them all}.
\newblock \bibinfo{journal}{\emph{arXiv preprint arXiv:2305.16355}} (\bibinfo{year}{2023}).
\newblock


\bibitem[Sultani et~al\mbox{.}(2018)]%
        {sultani2018real}
\bibfield{author}{\bibinfo{person}{Waqas Sultani}, \bibinfo{person}{Chen Chen}, {and} \bibinfo{person}{Mubarak Shah}.} \bibinfo{year}{2018}\natexlab{}.
\newblock \showarticletitle{Real-world anomaly detection in surveillance videos}. In \bibinfo{booktitle}{\emph{CVPR}}.
\newblock


\bibitem[Sun et~al\mbox{.}(2023)]%
        {sun2023eva}
\bibfield{author}{\bibinfo{person}{Quan Sun}, \bibinfo{person}{Yuxin Fang}, \bibinfo{person}{Ledell Wu}, \bibinfo{person}{Xinlong Wang}, {and} \bibinfo{person}{Yue Cao}.} \bibinfo{year}{2023}\natexlab{}.
\newblock \showarticletitle{Eva-clip: Improved training techniques for clip at scale}.
\newblock \bibinfo{journal}{\emph{arXiv preprint arXiv:2303.15389}} (\bibinfo{year}{2023}).
\newblock


\bibitem[Teed and Deng(2020)]%
        {teed2020raft}
\bibfield{author}{\bibinfo{person}{Zachary Teed} {and} \bibinfo{person}{Jia Deng}.} \bibinfo{year}{2020}\natexlab{}.
\newblock \showarticletitle{Raft: Recurrent all-pairs field transforms for optical flow}. In \bibinfo{booktitle}{\emph{Computer Vision--ECCV 2020: 16th European Conference, Glasgow, UK, August 23--28, 2020, Proceedings, Part II 16}}. Springer, \bibinfo{pages}{402--419}.
\newblock


\bibitem[Thakare et~al\mbox{.}(2023a)]%
        {thakare2023rareanom}
\bibfield{author}{\bibinfo{person}{Kamalakar~Vijay Thakare}, \bibinfo{person}{Debi~Prosad Dogra}, \bibinfo{person}{Heeseung Choi}, \bibinfo{person}{Haksub Kim}, {and} \bibinfo{person}{Ig-Jae Kim}.} \bibinfo{year}{2023}\natexlab{a}.
\newblock \showarticletitle{RareAnom: A Benchmark Video Dataset for Rare Type Anomalies}.
\newblock \bibinfo{journal}{\emph{Pattern Recognition}} (\bibinfo{year}{2023}).
\newblock


\bibitem[Thakare et~al\mbox{.}(2023b)]%
        {thakare2023dyannet}
\bibfield{author}{\bibinfo{person}{Kamalakar~Vijay Thakare}, \bibinfo{person}{Yash Raghuwanshi}, \bibinfo{person}{Debi~Prosad Dogra}, \bibinfo{person}{Heeseung Choi}, {and} \bibinfo{person}{Ig-Jae Kim}.} \bibinfo{year}{2023}\natexlab{b}.
\newblock \showarticletitle{DyAnNet: A Scene Dynamicity Guided Self-Trained Video Anomaly Detection Network}. In \bibinfo{booktitle}{\emph{WACV}}.
\newblock


\bibitem[Tian et~al\mbox{.}(2021)]%
        {tian2021weakly}
\bibfield{author}{\bibinfo{person}{Yu Tian}, \bibinfo{person}{Guansong Pang}, \bibinfo{person}{Yuanhong Chen}, \bibinfo{person}{Rajvinder Singh}, \bibinfo{person}{Johan~W Verjans}, {and} \bibinfo{person}{Gustavo Carneiro}.} \bibinfo{year}{2021}\natexlab{}.
\newblock \showarticletitle{Weakly-supervised video anomaly detection with robust temporal feature magnitude learning}. In \bibinfo{booktitle}{\emph{ICCV}}.
\newblock


\bibitem[Touvron et~al\mbox{.}(2023)]%
        {touvron2023llama}
\bibfield{author}{\bibinfo{person}{Hugo Touvron}, \bibinfo{person}{Thibaut Lavril}, \bibinfo{person}{Gautier Izacard}, \bibinfo{person}{Xavier Martinet}, \bibinfo{person}{Marie-Anne Lachaux}, \bibinfo{person}{Timoth{\'e}e Lacroix}, \bibinfo{person}{Baptiste Rozi{\`e}re}, \bibinfo{person}{Naman Goyal}, \bibinfo{person}{Eric Hambro}, \bibinfo{person}{Faisal Azhar}, {et~al\mbox{.}}} \bibinfo{year}{2023}\natexlab{}.
\newblock \showarticletitle{Llama: Open and efficient foundation language models}.
\newblock \bibinfo{journal}{\emph{arXiv}} (\bibinfo{year}{2023}).
\newblock


\bibitem[Tur et~al\mbox{.}(2023a)]%
        {tur2023exploring}
\bibfield{author}{\bibinfo{person}{Anil~Osman Tur}, \bibinfo{person}{Nicola Dall’Asen}, \bibinfo{person}{Cigdem Beyan}, {and} \bibinfo{person}{Elisa Ricci}.} \bibinfo{year}{2023}\natexlab{a}.
\newblock \showarticletitle{Exploring diffusion models for unsupervised video anomaly detection}. In \bibinfo{booktitle}{\emph{ICIP}}.
\newblock


\bibitem[Tur et~al\mbox{.}(2023b)]%
        {tur2023unsupervised}
\bibfield{author}{\bibinfo{person}{Anil~Osman Tur}, \bibinfo{person}{Nicola Dall’Asen}, \bibinfo{person}{Cigdem Beyan}, {and} \bibinfo{person}{Elisa Ricci}.} \bibinfo{year}{2023}\natexlab{b}.
\newblock \showarticletitle{Unsupervised Video Anomaly Detection with Diffusion Models Conditioned on Compact Motion Representations}. In \bibinfo{booktitle}{\emph{ICIAP}}.
\newblock


\bibitem[Vaswani et~al\mbox{.}(2017)]%
        {vaswani2017attention}
\bibfield{author}{\bibinfo{person}{Ashish Vaswani}, \bibinfo{person}{Noam Shazeer}, \bibinfo{person}{Niki Parmar}, \bibinfo{person}{Jakob Uszkoreit}, \bibinfo{person}{Llion Jones}, \bibinfo{person}{Aidan~N Gomez}, \bibinfo{person}{{\L}ukasz Kaiser}, {and} \bibinfo{person}{Illia Polosukhin}.} \bibinfo{year}{2017}\natexlab{}.
\newblock \showarticletitle{Attention is all you need}.
\newblock \bibinfo{journal}{\emph{NeurIPS}} (\bibinfo{year}{2017}).
\newblock


\bibitem[Veli{\v{c}}kovi{\'c} et~al\mbox{.}(2018)]%
        {velivckovic2018graph}
\bibfield{author}{\bibinfo{person}{Petar Veli{\v{c}}kovi{\'c}}, \bibinfo{person}{Guillem Cucurull}, \bibinfo{person}{Arantxa Casanova}, \bibinfo{person}{Adriana Romero}, \bibinfo{person}{Pietro Li{\`o}}, {and} \bibinfo{person}{Yoshua Bengio}.} \bibinfo{year}{2018}\natexlab{}.
\newblock \showarticletitle{Graph Attention Networks}. In \bibinfo{booktitle}{\emph{International Conference on Learning Representations}}.
\newblock


\bibitem[Wang and Cherian(2019)]%
        {wang2019gods}
\bibfield{author}{\bibinfo{person}{Jue Wang} {and} \bibinfo{person}{Anoop Cherian}.} \bibinfo{year}{2019}\natexlab{}.
\newblock \showarticletitle{Gods: Generalized one-class discriminative subspaces for anomaly detection}. In \bibinfo{booktitle}{\emph{ICCV}}.
\newblock


\bibitem[Wang et~al\mbox{.}(2025)]%
        {wang2025unifying}
\bibfield{author}{\bibinfo{person}{Nan Wang}, \bibinfo{person}{Yuantao Chen}, \bibinfo{person}{Lixing Xiao}, \bibinfo{person}{Weiqing Xiao}, \bibinfo{person}{Bohan Li}, \bibinfo{person}{Zhaoxi Chen}, \bibinfo{person}{Chongjie Ye}, \bibinfo{person}{Shaocong Xu}, \bibinfo{person}{Saining Zhang}, \bibinfo{person}{Ziyang Yan}, {et~al\mbox{.}}} \bibinfo{year}{2025}\natexlab{}.
\newblock \showarticletitle{Unifying Appearance Codes and Bilateral Grids for Driving Scene Gaussian Splatting}.
\newblock \bibinfo{journal}{\emph{arXiv preprint arXiv:2506.05280}} (\bibinfo{year}{2025}).
\newblock


\bibitem[Wang et~al\mbox{.}(2024)]%
        {wang2024cogvlm}
\bibfield{author}{\bibinfo{person}{Weihan Wang}, \bibinfo{person}{Qingsong Lv}, \bibinfo{person}{Wenmeng Yu}, \bibinfo{person}{Wenyi Hong}, \bibinfo{person}{Ji Qi}, \bibinfo{person}{Yan Wang}, \bibinfo{person}{Junhui Ji}, \bibinfo{person}{Zhuoyi Yang}, \bibinfo{person}{Lei Zhao}, \bibinfo{person}{Song XiXuan}, {et~al\mbox{.}}} \bibinfo{year}{2024}\natexlab{}.
\newblock \showarticletitle{Cogvlm: Visual expert for pretrained language models}.
\newblock \bibinfo{journal}{\emph{Advances in Neural Information Processing Systems}}  \bibinfo{volume}{37} (\bibinfo{year}{2024}), \bibinfo{pages}{121475--121499}.
\newblock


\bibitem[Wu et~al\mbox{.}(2022)]%
        {wu2022self}
\bibfield{author}{\bibinfo{person}{Jhih-Ciang Wu}, \bibinfo{person}{He-Yen Hsieh}, \bibinfo{person}{Ding-Jie Chen}, \bibinfo{person}{Chiou-Shann Fuh}, {and} \bibinfo{person}{Tyng-Luh Liu}.} \bibinfo{year}{2022}\natexlab{}.
\newblock \showarticletitle{Self-supervised Sparse Representation for Video Anomaly Detection}. In \bibinfo{booktitle}{\emph{ECCV}}.
\newblock


\bibitem[Wu and Liu(2021)]%
        {wu2021learning}
\bibfield{author}{\bibinfo{person}{Peng Wu} {and} \bibinfo{person}{Jing Liu}.} \bibinfo{year}{2021}\natexlab{}.
\newblock \showarticletitle{Learning causal temporal relation and feature discrimination for anomaly detection}.
\newblock \bibinfo{journal}{\emph{IEEE TIP}} (\bibinfo{year}{2021}).
\newblock


\bibitem[Wu et~al\mbox{.}(2020)]%
        {wu2020not}
\bibfield{author}{\bibinfo{person}{Peng Wu}, \bibinfo{person}{Jing Liu}, \bibinfo{person}{Yujia Shi}, \bibinfo{person}{Yujia Sun}, \bibinfo{person}{Fangtao Shao}, \bibinfo{person}{Zhaoyang Wu}, {and} \bibinfo{person}{Zhiwei Yang}.} \bibinfo{year}{2020}\natexlab{}.
\newblock \showarticletitle{Not only look, but also listen: Learning multimodal violence detection under weak supervision}. In \bibinfo{booktitle}{\emph{ECCV}}.
\newblock


\bibitem[Yan et~al\mbox{.}(2024a)]%
        {yan2024renderworld}
\bibfield{author}{\bibinfo{person}{Ziyang Yan}, \bibinfo{person}{Wenzhen Dong}, \bibinfo{person}{Yihua Shao}, \bibinfo{person}{Yuhang Lu}, \bibinfo{person}{Liu Haiyang}, \bibinfo{person}{Jingwen Liu}, \bibinfo{person}{Haozhe Wang}, \bibinfo{person}{Zhe Wang}, \bibinfo{person}{Yan Wang}, \bibinfo{person}{Fabio Remondino}, {et~al\mbox{.}}} \bibinfo{year}{2024}\natexlab{a}.
\newblock \showarticletitle{Renderworld: World model with self-supervised 3d label}.
\newblock \bibinfo{journal}{\emph{arXiv preprint arXiv:2409.11356}} (\bibinfo{year}{2024}).
\newblock


\bibitem[Yan et~al\mbox{.}(2024b)]%
        {yan20243dsceneeditor}
\bibfield{author}{\bibinfo{person}{Ziyang Yan}, \bibinfo{person}{Lei Li}, \bibinfo{person}{Yihua Shao}, \bibinfo{person}{Siyu Chen}, \bibinfo{person}{Zongkai Wu}, \bibinfo{person}{Jenq-Neng Hwang}, \bibinfo{person}{Hao Zhao}, {and} \bibinfo{person}{Fabio Remondino}.} \bibinfo{year}{2024}\natexlab{b}.
\newblock \showarticletitle{3dsceneeditor: Controllable 3d scene editing with gaussian splatting}.
\newblock \bibinfo{journal}{\emph{arXiv preprint arXiv:2412.01583}} (\bibinfo{year}{2024}).
\newblock


\bibitem[Yan et~al\mbox{.}(2023)]%
        {yan2023nerfbk}
\bibfield{author}{\bibinfo{person}{Ziyang Yan}, \bibinfo{person}{Gabriele Mazzacca}, \bibinfo{person}{Simone Rigon}, \bibinfo{person}{Elisa~Mariarosaria Farella}, \bibinfo{person}{Pawel Trybala}, \bibinfo{person}{Fabio Remondino}, {et~al\mbox{.}}} \bibinfo{year}{2023}\natexlab{}.
\newblock \showarticletitle{NeRFBK: a holistic dataset for benchmarking NeRF-based 3D reconstruction}.
\newblock \bibinfo{journal}{\emph{International Archives of the Photogrammetry, Remote Sensing and Spatial Information Sciences}} \bibinfo{volume}{48}, \bibinfo{number}{1} (\bibinfo{year}{2023}), \bibinfo{pages}{219--226}.
\newblock


\bibitem[Yan et~al\mbox{.}(2025)]%
        {yan2025learning}
\bibfield{author}{\bibinfo{person}{Ziyang Yan}, \bibinfo{person}{Nazanin Padkan}, \bibinfo{person}{Pawe{\l} Tryba{\l}a}, \bibinfo{person}{Elisa~Mariarosaria Farella}, {and} \bibinfo{person}{Fabio Remondino}.} \bibinfo{year}{2025}\natexlab{}.
\newblock \showarticletitle{Learning-Based 3D Reconstruction Methods for Non-Collaborative Surfaces—A Metrological Evaluation}.
\newblock \bibinfo{journal}{\emph{Metrology}} \bibinfo{volume}{5}, \bibinfo{number}{2} (\bibinfo{year}{2025}), \bibinfo{pages}{20}.
\newblock


\bibitem[Ye et~al\mbox{.}(2024)]%
        {ye2024mplug}
\bibfield{author}{\bibinfo{person}{Qinghao Ye}, \bibinfo{person}{Haiyang Xu}, \bibinfo{person}{Jiabo Ye}, \bibinfo{person}{Ming Yan}, \bibinfo{person}{Anwen Hu}, \bibinfo{person}{Haowei Liu}, \bibinfo{person}{Qi Qian}, \bibinfo{person}{Ji Zhang}, {and} \bibinfo{person}{Fei Huang}.} \bibinfo{year}{2024}\natexlab{}.
\newblock \showarticletitle{mplug-owl2: Revolutionizing multi-modal large language model with modality collaboration}. In \bibinfo{booktitle}{\emph{Proceedings of the ieee/cvf conference on computer vision and pattern recognition}}. \bibinfo{pages}{13040--13051}.
\newblock


\bibitem[Zaheer et~al\mbox{.}(2020)]%
        {zaheer2020claws}
\bibfield{author}{\bibinfo{person}{Muhammad~Zaigham Zaheer}, \bibinfo{person}{Arif Mahmood}, \bibinfo{person}{Marcella Astrid}, {and} \bibinfo{person}{Seung-Ik Lee}.} \bibinfo{year}{2020}\natexlab{}.
\newblock \showarticletitle{Claws: Clustering assisted weakly supervised learning with normalcy suppression for anomalous event detection}. In \bibinfo{booktitle}{\emph{ECCV}}.
\newblock


\bibitem[Zaheer et~al\mbox{.}(2022)]%
        {zaheer2022generative}
\bibfield{author}{\bibinfo{person}{M~Zaigham Zaheer}, \bibinfo{person}{Arif Mahmood}, \bibinfo{person}{M~Haris Khan}, \bibinfo{person}{Mattia Segu}, \bibinfo{person}{Fisher Yu}, {and} \bibinfo{person}{Seung-Ik Lee}.} \bibinfo{year}{2022}\natexlab{}.
\newblock \showarticletitle{Generative cooperative learning for unsupervised video anomaly detection}. In \bibinfo{booktitle}{\emph{CVPR}}.
\newblock


\bibitem[Zanella et~al\mbox{.}(2024)]%
        {zanella2024harnessing}
\bibfield{author}{\bibinfo{person}{Luca Zanella}, \bibinfo{person}{Willi Menapace}, \bibinfo{person}{Massimiliano Mancini}, \bibinfo{person}{Yiming Wang}, {and} \bibinfo{person}{Elisa Ricci}.} \bibinfo{year}{2024}\natexlab{}.
\newblock \showarticletitle{Harnessing large language models for training-free video anomaly detection}. In \bibinfo{booktitle}{\emph{Proceedings of the IEEE/CVF Conference on Computer Vision and Pattern Recognition}}. \bibinfo{pages}{18527--18536}.
\newblock


\bibitem[Zeng et~al\mbox{.}(2025b)]%
        {zeng2025local}
\bibfield{author}{\bibinfo{person}{Fanhu Zeng}, \bibinfo{person}{Zhen Cheng}, \bibinfo{person}{Fei Zhu}, \bibinfo{person}{Hongxin Wei}, {and} \bibinfo{person}{Xu-Yao Zhang}.} \bibinfo{year}{2025}\natexlab{b}.
\newblock \showarticletitle{Local-Prompt: Extensible Local Prompts for Few-Shot Out-of-Distribution Detection}. In \bibinfo{booktitle}{\emph{The Thirteenth International Conference on Learning Representations}}.
\newblock


\bibitem[Zeng et~al\mbox{.}(2025a)]%
        {zeng2025towards}
\bibfield{author}{\bibinfo{person}{Fanhu Zeng}, \bibinfo{person}{Zhen Cheng}, \bibinfo{person}{Fei Zhu}, {and} \bibinfo{person}{Xu-Yao Zhang}.} \bibinfo{year}{2025}\natexlab{a}.
\newblock \showarticletitle{Towards Efficient and General-Purpose Few-Shot Misclassification Detection for Vision-Language Models}.
\newblock \bibinfo{journal}{\emph{arXiv preprint arXiv:2503.20492}} (\bibinfo{year}{2025}).
\newblock


\bibitem[Zeng et~al\mbox{.}(2025c)]%
        {zeng2025parameter}
\bibfield{author}{\bibinfo{person}{Fanhu Zeng}, \bibinfo{person}{Haiyang Guo}, \bibinfo{person}{Fei Zhu}, \bibinfo{person}{Li Shen}, {and} \bibinfo{person}{Hao Tang}.} \bibinfo{year}{2025}\natexlab{c}.
\newblock \showarticletitle{Parameter Efficient Merging for Multimodal Large Language Models with Complementary Parameter Adaptation}.
\newblock \bibinfo{journal}{\emph{arXiv preprint arXiv:2502.17159}} (\bibinfo{year}{2025}).
\newblock


\bibitem[Zeng et~al\mbox{.}(2025d)]%
        {zeng2025mambaic}
\bibfield{author}{\bibinfo{person}{Fanhu Zeng}, \bibinfo{person}{Hao Tang}, \bibinfo{person}{Yihua Shao}, \bibinfo{person}{Siyu Chen}, \bibinfo{person}{Ling Shao}, {and} \bibinfo{person}{Yan Wang}.} \bibinfo{year}{2025}\natexlab{d}.
\newblock \showarticletitle{MambaIC: State Space Models for High-Performance Learned Image Compression}.
\newblock \bibinfo{journal}{\emph{arXiv preprint arXiv:2503.12461}} (\bibinfo{year}{2025}).
\newblock


\bibitem[Zeng and Yu(2024)]%
        {zeng2024m2m}
\bibfield{author}{\bibinfo{person}{Fanhu Zeng} {and} \bibinfo{person}{Deli Yu}.} \bibinfo{year}{2024}\natexlab{}.
\newblock \showarticletitle{M2M-TAG: Training-Free Many-to-Many Token Aggregation for Vision Transformer Acceleration}. In \bibinfo{booktitle}{\emph{Workshop on Machine Learning and Compression, NeurIPS 2024}}.
\newblock


\bibitem[Zeng et~al\mbox{.}(2024)]%
        {zeng2024modalprompt}
\bibfield{author}{\bibinfo{person}{Fanhu Zeng}, \bibinfo{person}{Fei Zhu}, \bibinfo{person}{Haiyang Guo}, \bibinfo{person}{Xu-Yao Zhang}, {and} \bibinfo{person}{Cheng-Lin Liu}.} \bibinfo{year}{2024}\natexlab{}.
\newblock \showarticletitle{Modalprompt: Dual-modality guided prompt for continual learning of large multimodal models}.
\newblock \bibinfo{journal}{\emph{arXiv preprint arXiv:2410.05849}} (\bibinfo{year}{2024}).
\newblock


\bibitem[Zhang et~al\mbox{.}(2023)]%
        {zhang2023video}
\bibfield{author}{\bibinfo{person}{Hang Zhang}, \bibinfo{person}{Xin Li}, {and} \bibinfo{person}{Lidong Bing}.} \bibinfo{year}{2023}\natexlab{}.
\newblock \showarticletitle{Video-llama: An instruction-tuned audio-visual language model for video understanding}.
\newblock \bibinfo{journal}{\emph{arXiv preprint arXiv:2306.02858}} (\bibinfo{year}{2023}).
\newblock


\bibitem[Zhu et~al\mbox{.}(2023)]%
        {zhu2023languagebind}
\bibfield{author}{\bibinfo{person}{Bin Zhu}, \bibinfo{person}{Bin Lin}, \bibinfo{person}{Munan Ning}, \bibinfo{person}{Yang Yan}, \bibinfo{person}{Jiaxi Cui}, \bibinfo{person}{HongFa Wang}, \bibinfo{person}{Yatian Pang}, \bibinfo{person}{Wenhao Jiang}, \bibinfo{person}{Junwu Zhang}, \bibinfo{person}{Zongwei Li}, {et~al\mbox{.}}} \bibinfo{year}{2023}\natexlab{}.
\newblock \showarticletitle{Languagebind: Extending video-language pretraining to n-modality by language-based semantic alignment}.
\newblock \bibinfo{journal}{\emph{arXiv preprint arXiv:2310.01852}} (\bibinfo{year}{2023}).
\newblock


\end{thebibliography}

\end{CJK}
\end{document}